\documentclass[final,5p,times,twocolumn]{elsarticle}

\usepackage{amssymb}
\usepackage{amsfonts}
\usepackage{amsthm}
\usepackage{graphics} % for pdf, bitmapped graphics files
\usepackage{epsfig} % for postscript graphics files
\usepackage{amsmath} % assumes amsmath package installed
\usepackage{subfigure}
\usepackage{multirow}
\usepackage{makecell}
\usepackage{multirow}
\usepackage{graphicx}
\usepackage{lscape}
\usepackage{multicol} %multi table
\usepackage{graphicx, epstopdf}
\usepackage[table,xcdraw]{xcolor}
\usepackage{caption}
\usepackage{xspace,epsfig,url}
\usepackage{enumerate}
\usepackage{float}
\usepackage{epsf}
\usepackage{psfrag}
\usepackage{verbatim}
\usepackage[all]{xy}
\usepackage{color,soul}
\usepackage{bm}
\usepackage{comment}
\usepackage{cases}
\usepackage{hyperref}
\usepackage{physics}
\usepackage{amssymb}% http://ctan.org/pkg/amssymb
\usepackage{pifont}% http://ctan.org/pkg/pifont
\usepackage{array} %multi table
\usepackage[ruled,vlined,linesnumbered]{algorithm2e}

\DeclareMathOperator{\sign}{sign}

\journal{ArXiv}

\begin{document}

\begin{frontmatter}

\title{Design Optimizer for Soft Growing Robot Manipulators\\in Three-Dimensional Environments}

\author[inst1]{Ahmet Astar}
\author[inst1]{Ozan Nurcan}
\author[inst1]{Erk Demirel}
\author[inst1]{Emir Ozen}
\author[inst1]{Ozan Kutlar}
\author[inst1,*]{Fabio Stroppa}

\affiliation[inst1]{organization={Computer Engineering Department, Kadir Has University},
            addressline={Cibali, Kadir Has Cd., Fatih}, 
            city={Istanbul},
            postcode={34083}, 
            country={Turkey}}
\affiliation[*]{Corresponding Author: fabio.stroppa@khas.edu.tr}

\begin{abstract}
%intro
Soft growing robots are novel devices that mimic plant-like growth for navigation in cluttered or dangerous environments. Their ability to adapt to surroundings, combined with advancements in actuation and manufacturing technologies, allows them to perform specialized manipulation tasks. 
%proposal
This work presents an approach for design optimization of soft growing robots; specifically, the three-dimensional extension of the optimizer designed for planar manipulators. This tool is intended to be used by engineers and robot enthusiasts before manufacturing their robot: it suggests the optimal size of the robot for solving a specific task.
The design process models a multi-objective optimization problem to refine a soft manipulator's kinematic chain. Thanks to the novel Rank Partitioning algorithm integrated into Evolutionary Computation (EC) algorithms, this method achieves high precision in reaching targets and is efficient in resource usage. 
%results
Results show significantly high performance in solving three-dimensional tasks, whereas comparative experiments indicate that the optimizer features robust output when tested with different EC algorithms, particularly genetic algorithms.
\end{abstract}


\end{frontmatter}



\section{Introduction}
\label{sec:intro}

\begin{figure}[t!]
	\centering
        \subfigure[\protect\url{}\label{fig:soft_growing_robot_drawing}]%
	{\includegraphics[height=4cm]{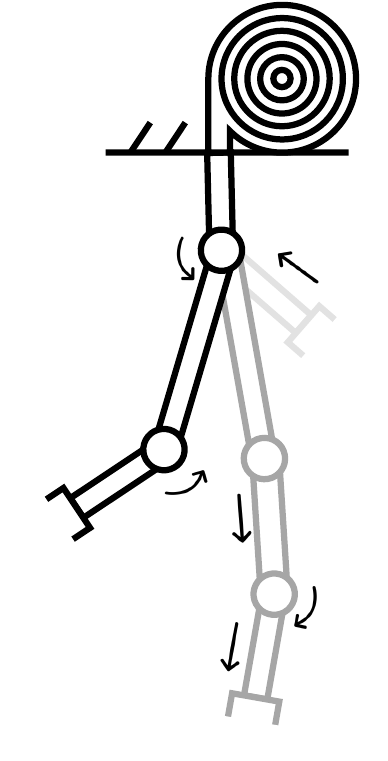}}\hfill
        \subfigure[\protect\url{}\label{fig:maze}]%
	{\includegraphics[height=4cm]{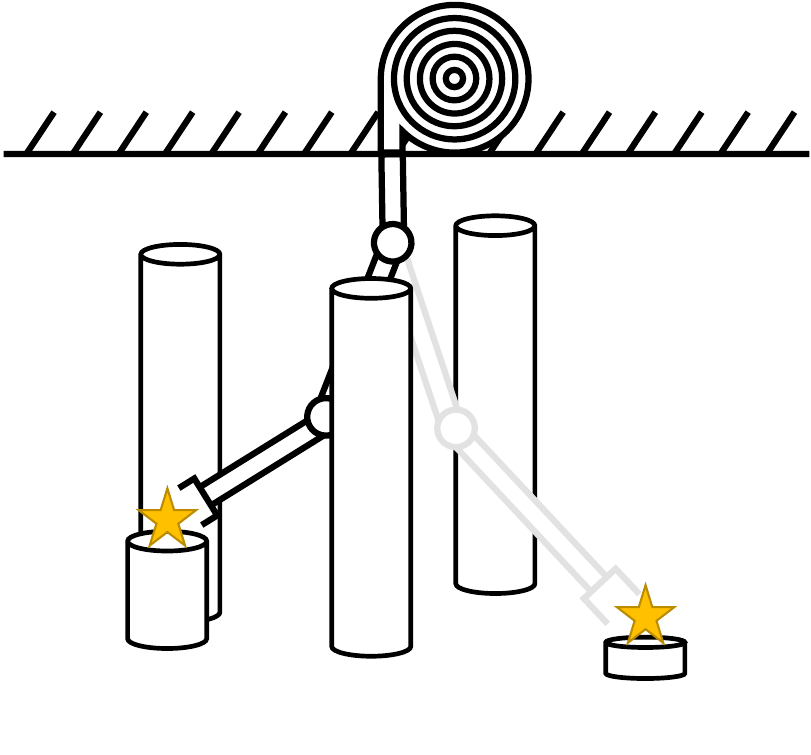}}\hfill
        \subfigure[\protect\url{}\label{fig:inflated_robot}]%
        {\includegraphics[height=4cm]{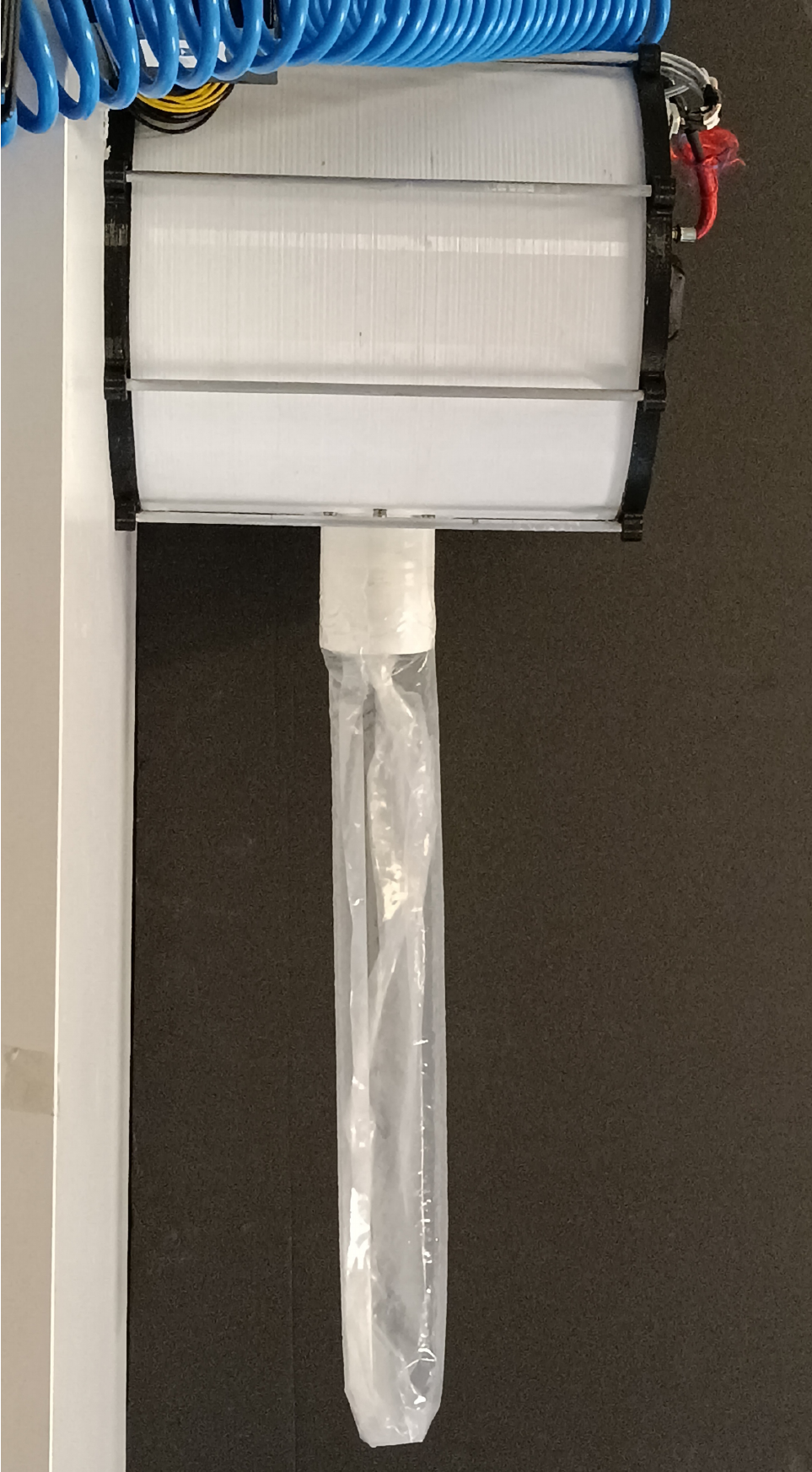}}
	\caption{(a) A soft growing manipulator with continuum links and variable discrete joints~\cite{do2024stiffness}. (b) Sketch of a three-dimensional soft growing robot reaching two targets in an environment with obstacles (side view). (c) A picture of a physical manipulator.}
	\label{fig:example}
\end{figure}

% 1. The Hook
% Sentence stating what the domain is and why it is important}
% What is the overall problem or situation in that domain.}
Soft robotics is a relatively new field of intelligent mechanical systems that aims at overcoming some of the biggest limitations of their rigid counterparts~\cite{whitesides2018soft}. Rigid robots (i) may struggle to interact with delicate objects or environments, such as human tissues, glass objects, or sensitive components in archaeological sites; (ii) have intricate designs and their materials often result in high costs; and (iii) due to their strength and stiffness, the safety of human operators can be a concern in a human-machine interaction scenario~\cite{ogorodnikova2009safe}. On the other hand, soft robots are inspired by living organisms, with designs exploiting the natural compliance of their materials to facilitate smooth and intuitive physical interactions with the environment~\cite{webster2010design, rus2015design, whitesides2018soft, hawkes2021hard}. Moreover, their materials and actuation methods are typically safer, cheaper, and simpler to manufacture or assemble compared to conventional rigid robots~\cite{abidi2017intrinsic}.
Additionally, soft robots have the ability to extend or retract through a process known as \textit{eversion} -- i.e., like a linear actuator, but with the advantage of occupying no space when completely retracted because the soft material can be folded. This unique degree of freedom (DoF) is inspired by the growth patterns of plants, leading to these robots often being referred to as \textit{vine} robots~\cite{coad2019vine, hawkes2017soft, blumenschein2017modeling} -- or more generically soft \textit{growing} robots. Several studies have demonstrated their effectiveness in manipulation tasks that require handling of payloads~\cite{ansari2017towards, stroppa2020human, stroppa2023shared,allen2024modeling}. A major innovation was introduced by Do et al.~\cite{do2024stiffness} with continuum links and discrete joints: thanks to dynamic layer jamming, the robot can independently stiffen the different parts of its body, resulting in a joint. As illustrated in Fig.~\ref{fig:soft_growing_robot_drawing}, this allows soft growing robots to be modeled with the classic kinematic chain used for their rigid counterparts, with the substantial difference that each link also grows and retracts.

% 2. The Gap
% What are the conclusions of existing literature for that problem/situation in this domain?
% What is the problem or issue with that existing literature?

Soft robots are considerably less costly than their rigid counterparts, and this makes them accessible to everyone, including amateur roboticists. Every robotic enthusiast could create their soft growing manipulator for one or more well-defined specific tasks~\cite{schulz2017interactive, morimoto2018toward, exarchos2022task}. However, specifying and integrating the components needed for a task-specific robot remains a complex challenge for human designers. With the capability of growing and retracting, the robot's tip (namely the \textit{end effector} or where the gripper/manipulative device is placed) can move anywhere along its kinematic chain. Consequently, their kinematic modeling and actuation pose significant difficulties. Preferably, a designer only needs to define the task itself, including any hardware or environmental constraint, and use an optimizer to retrieve optimal designs~\cite{hiller2011automatic}.

Determining a soft growing robot's design is a complex engineering problem that mathematical optimization can easily address~\cite{stroppa2024optimizing}. A \textit{design} defines the parameters and attributes describing a soft growing robot, ultimately guiding its manufacturing process. This includes aspects such as link lengths, joint positions and their range of motion (if applicable), materials used, and the overall shape or type of the robot. 
This problem has been addressed with several optimization methods: the interior point algorithm~\cite{tan2017simultaneous,lloyd2020optimal,kim2014optimizing,ros2019design,usevitch2020untethered,ghoreishi2021bayesian}, the Levenberg-Marquardt algorithm~\cite{tan2017simultaneous,lai2022constrained}, simplex methods~\cite{abbaszadeh2022design,burgner2013computational,bergeles2015concentric,rucker2011statics,rucker2010geometrically,rucker2010model}, the steepest descent method~\cite{chen2021enhancing,chen2018topology}, linear approximations~\cite{schiller2020lightweight}, cyclic coordinate descent~\cite{zhang2021soft}, Bayesian optimization~\cite{ghoreishi2021bayesian}, the method of moving asymptotes~\cite{wang2020topology,li2023tailoring}, the Newton-Raphson method~\cite{morgan2023towards}, greedy algorithms~\cite{koehler2020model}, hill climbing~\cite{fathurrohim2022maximizing}, simulated annealing~\cite{ghoreishi2021bayesian}, pattern search~\cite{bedell2011design,anor2011algorithms}, specific heuristics~\cite{wu2022novel,wu2022crrik}, the parametric quadratic programming algorithm~\cite{adagolodjo2021coupling,coevoet2017optimization}, and the Broyden-Fletcher-Goldfarb-Shanno algorithm~\cite{maloisel2023optimal}. 
However, these traditional optimization methods often struggle with navigating highly complex, nonlinear, and multimodal design spaces, making it difficult to find global optima in engineering design problems.
That is why engineering problems often rely on artificial intelligence, specifically the subfield of Evolutionary Computation (EC), a sub-field of Artificial Intelligence~\cite{russell2010artificial}. EC algorithms, which are natural-inspired metaheuristic optimizers, are widely used in mechanical and robotic design~\cite{coello1998using,iwasaki2000evolutionary,lim2005inverse,filipiak2015infeasibility,rokbani2022beta,stroppa2023optimizing}. Specifically for soft robotics, many studies have used EC algorithms to solve their design optimization problems: genetic algorithms for single~\cite{rieffel2009evolving,hiller2011automatic,dinakaran2023performa,sui2022task,berger2015growing,marzougui2022comparative,abbaszadeh2022design,medvet2021biodiversity,fathurrohim2022maximizing} and multi-objective optimization~\cite{liu2022simulation,fitzgerald2021evolving,fitzgerald2022evolving,kimura2021modularized,bodily2017multi,stroppa2024design}, particle swarm optimization~\cite{abbaszadeh2022design,djeffal2021kinematics,atia2022reconfigurable,chen2020obstacle,gao2017kinematics}, differential evolution~\cite{tan2017simultaneous}, age-fitness-pareto optimization~\cite{kriegman2017minimal}, the estimation of distribution algorithm~\cite{cheong2021optimal}, covariance matrix adaptation evolution strategy~\cite{medvet2021biodiversity,ferigo2021beyond,ferigo2022optimizing}, and neuroevolution~\cite{kimura2021modularized,cheney2015evolving}.
These studies aim to optimize a soft robot's design by finding the simplest possible configuration (i.e., the minimum number of sections) capable of performing a specific task or set of tasks~\cite{anor2011algorithms}. The design optimization typically requires solving the robot's kinematics, which, in the case of a soft manipulator, also involves finding the correct curvature along the body~\cite{chen2020obstacle}.

Fig.~\ref{fig:maze} shows an example of a soft growing manipulator with discrete joints~\cite{do2024stiffness} specifically designed to solve a given manipulation task; whereas Fig.~\ref{fig:inflated_robot} shows a real picture of the manipulator. This specific problem has been addressed in the past by Exarchos et al.~\cite{exarchos2022task} in a two-dimensional scenario only -- i.e., the manipulator acts in a planar environment. The same problem was addressed by our previous work~\cite{stroppa2024design}, outperforming the state-of-the-art in terms of precision in reaching targets, smoothness of the configurations (i.e., non-wiggly or wavy movements to reach targets), and overall length of manipulators.

% 3. The Study
% Indicate how this study addresses that problem or issue (Must start with "In this work, we...")
% Describe the study, sample, and method for addressing that problem or issue
In this work, we extend our previous design optimizer for planar robots~\cite{stroppa2024design} into the third dimension to address more realistic scenarios with robots acting in a complex space -- i.e., from two dimensions (2D) to three (3D). The design is modeled as a Multi-Objective Optimization Problem (MOOP) and addressed with EC algorithms. Specifically, we exploit the Rank Partitioning algorithm (introduced in the previous work~\cite{stroppa2024design}) to solve the MOOP without the need to retrieve the entire Pareto optimal front~\cite{deb1999multi,miettinen2012nonlinear}. This allows the optimizer to focus only on specific desired settings, thus minimizing the disadvantage of having trade-off designs. 
%Rank partitioning similar to~\cite{goldman2014parameter,rafiq2021pyramid}.

% 4. Contribution
% Describe what you found
% State explicitly how these findings extend and contribute to existing knowledge
The contributions of this work are: (i) an extension of the innovative AI-based tool for robotic engineers presented in our previous work~\cite{stroppa2024design} by introducing the third dimension in the design; (ii) a mathematical formulation for this design optimization problem based on the kinematics proposed by Do et al.~\cite{do2024stiffness}, which to the best of our knowledge has not been addressed in 3D before; (iii) the inclusion of additional constraints to the optimization problem that were overlooked or not needed in the 2D case; and (iv) a comparative robustness analysis of the optimizer and the Rank Partitioning algorithm when applied to four different EC algorithms.

% 5. Outline/Roadmap
The rest of the work is divided as follows: Section~\ref{sec:background} describes the specific soft growing manipulator and the scenario we addressed; 
Section~\ref{sec:optimization_problem} offers a comprehensive overview of the optimization problem and a detailed walkthrough of the proposed solution utilizing EC algorithms; Section~\ref{sec:experiments} presents an experiment to test the optimality of the method, particularly comparing how four different EC algorithms perform with Rank Partitioning; and Section~\ref{sec:conclusion} concludes the work illustrating possible future directions.

\begin{figure}[b!]
	\centering
        {\includegraphics[height=6cm]{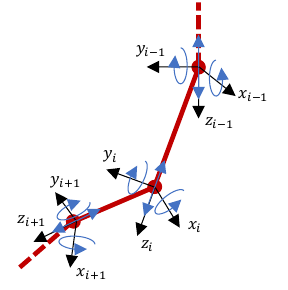}}
	\caption{Degrees of freedom (blue arrows) of the soft growing manipulator addressed in the current work, originally designed by Do et al.~\cite{do2024stiffness}. Each joint can rotate on x and y-axes, growing or retracting on the z-axis. Each joint has its own reference frame, whose z-axis is aligned in the direction of the previous link. }
	\label{fig:dof}
\end{figure}

\section{Soft Growing Manipulator for the Proposed Scenario}
\label{sec:background}

The soft growing manipulator examined in this study is based on the design proposed by Do et al.~\cite{do2024stiffness}, which combines characteristics of both rigid and soft robots -- Fig.~\ref{fig:dof} depicts a portion of the manipulator's model to show its DoFs. The manipulator consists of links that can be individually adjusted to stiffen or stay flexible. When a link is not stiffened, it can be deliberately steered and maneuvered using cables to function as a revolute joint. When a link is stiffened, pulling its respective cable will not change its orientation; however, the following link (if not stiffeners) will be steered in the desired direction. Thanks to this technology, each link can steer independently by using only two cables for each direction (i.e., clockwise and anti-clockwise rotation around the x-axis, clockwise and anti-clockwise rotation around the y-axis). Lastly, growth and retraction (eversion) are made possible by the structure of the robot's body, which is an inflatable beam composed of a thin-walled, compliant material that everts pneumatically. Starting in a fully folded state, pressurizing the interior with air unfolding the tip, pulling new material through its body~\cite{hawkes2017soft, mishima2003development, tsukagoshi2011tip}.

\begin{figure}[t! ]
    \centering
    {\includegraphics[width=7cm]{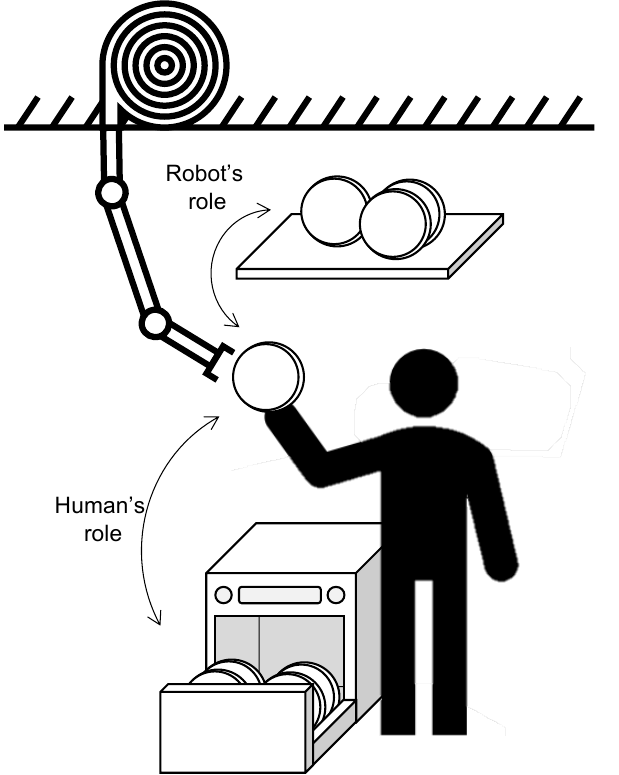}}
    \caption{Example of human-machine interaction with a three-dimensional soft growing robot. A soft manipulator mounted on the ceiling reaches down to help a human load/unload a dishwasher. The human's role is to take dishes from the machine, whereas the robot's role is to safely place them on the shelf.}
	\label{fig:case_hmi}
\end{figure}

This type of manipulator can be used in manipulation tasks~\cite{exarchos2022task} when equipped with a gripper at its tip (i.e., the end effector). This is usually done with a magnetic tip mount that moves as the robot grows, sliding on the body material, remaining anchored to the tip~\cite{jeong2020tip,coad2020retraction,stroppa2020human,stroppa2023shared}. In the specific scenario analyzed by this work, the robot is hanged on the ceiling and grows in favor of gravity to reach items and targets placed in the environment -- much like a claw machine at the fair. Fig.~\ref{fig:case_hmi} shows a possible real-case example in which the robot assists a human while unloading a dishwasher: the human selects the item from the dishwasher, hands it to the robot, and the robot places it in the designated spot on a shelf -- a classic example of human-machine interaction. 
In this scenario, a soft robot is inherently safer than a rigid robot due to its flexibility and compliance~\cite{abidi2017intrinsic}. Unlike rigid robots, which can exert high forces and have hard surfaces, soft robots are made of materials that can deform upon contact. This reduces the risk of injury if a collision with a human occurs. Additionally, the adaptable nature of soft robots allows them to handle objects of various shapes and fragilities with greater care, minimizing the chance of damaging items while also making the interaction more intuitive and safer for the user.

\begin{figure*}[h!]
    \centering
    \subfigure[\protect\url{}\label{fig:problem_example_a}Task]%
    {\includegraphics[width=5cm]{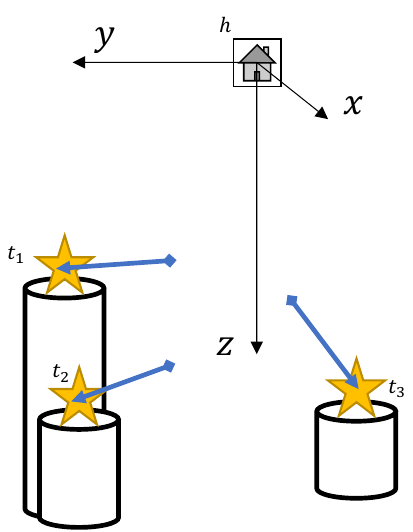}} \hspace{1cm}
    \subfigure[\protect\url{}\label{fig:problem_example_b}Solution]%
    {\includegraphics[width=5cm]{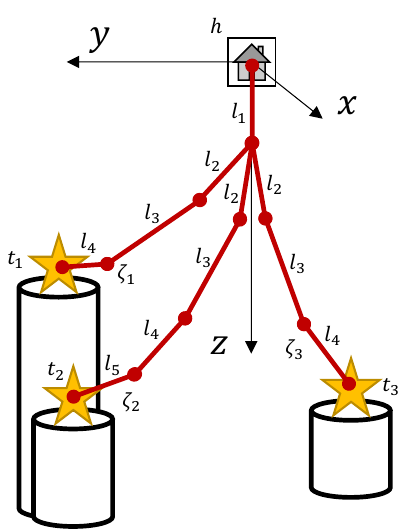}}
    \caption{(a) Task example: three targets placed on different pillars with a desired orientation. The robot grows from its base $h$ such that its final link is aligned with each target. This task has three targets ($|t|=3$) and three obstacles ($|o|=3$) -- i.e., the pillars where the targets are placed. (b) Optimizer's output example: the solution $\psi$ comprises three configurations $\zeta_{1\rightarrow3}$ that share a unique design $\delta$ (i.e., the three configurations $\zeta$ share the same link lengths). Because the robot may use only a limited amount of links to reach a specific target, depending on its distance to it, each configuration can have a different number of employed links. In this example, the number of links whose endpoint corresponds to the robot's tip are $\bar{n}_1=4$, $\bar{n}_2=5$, $\bar{n}_3=4$.}
    \label{fig:problem_example}
\end{figure*} 

\section{Optimization Problem}
\label{sec:optimization_problem}

This section describes the optimization problem, its formalization, and how the objectives were engineered to formulate a proper fitness evaluation. Since the details are similar to the 2D case, with the exception of adding a further dimension for the rotational joints, we briefly summarize this description and invite readers to find more details in our previous work~\cite{stroppa2024design}. However, unlike our previous work, more information has been added regarding the specifics of the procedures, which can be found in Section 1.1 of the Supplementary Material.

The problem requires us to find the best design for a soft growing robot manipulator to solve a specific task -- i.e., the set of link lengths composing the body of the robot such that the robot can easily reach every target in the environment. Our optimizer will retrieve the design automatically.
We defined this problem in both 2D and 3D based on the effect of gravity on the system: a robot navigating a tunnel or a pipe system can be thought of as 2D; but when the environment demands navigating different levels or layers, it requires 3D exploration. 
The 2D scenario was tackled in our previous work~\cite{stroppa2024design}, and the current work extends it to the third dimension. Fig.~\ref{fig:problem_example} shows a 3D representation of the problem, where the robot is mounted on the environment's ceiling and grows in favor of gravity -- oriented in the positive direction of the z-axis. To simplify the problem, we did not consider gravity in the model. Fig.~\ref{fig:problem_example_a} shows an example of a 3D task in which the manipulator needs to reach three targets according to a desired reaching orientation. The robot should then align its last link -- or more than one -- so that it lies on the requested straight line. One example of design is shown in Fig.~\ref{fig:problem_example_b}, with three configurations (one for each target) sharing the same link lengths except for the last one.

The robot begins its extension from its container, known as the home base $h$. Based on the home base's pose (position and orientation) and the environment, the optimizer determines the optimal robot design so that the manipulator can reach every target without violating any constraint (see Section~\ref{sec:constraints}). The joint number and placement define the parameters of the design, which must be known prior to the robot's manufacturing and assembly -- specifically, the design $\delta$ is a set of link lengths defining a single manipulator. 

This results in a single optimal solution $\psi$ having multiple configurations $\zeta$ for each target (i.e., a different manipulator's orientation reaching each target) sharing a unique link-length design $\delta$.
Configurations can differ from each other due to eversion: they may use a different link number because not all the links may actually be employed -- i.e., they are folded within the robot's body. Additionally, it is also not guaranteed that the last link will be completely everted, depending on how far the last employed joint is from the specific target.

We model this scenario as an optimization problem: beginning with a random solution (random link lengths and joint angles for each target/configuration) generated through forward kinematics, it minimizes the distance from the end effector and each of the targets' poses. 
The design (link lengths and, consequently, link angles) are retrieved as an inverse kinematic solver.

\begin{table}[t!]
\centering
\caption{Nomenclature of the terms defined in the Optimization Problem}
\label{tab:nomenclature}
\resizebox{0.9\columnwidth}{!}{%
%\vspace{0.2cm}
    \begin{tabular}{cc}
    \hline
    \multicolumn{1}{c}{\textbf{Symbol}} & \textbf{Description} \\ \hline
    $\delta$ & design \\ \hline
    $\psi$ & solution \\ \hline
    $\zeta$ & configuration \\ \hline
    $n$ & maximum number of links/nodes \\ \hline
    $\theta^{(x)}$ & joint rotation around x-axis \\ \hline
    $\theta^{(y)}$ & joint rotation around y-axis \\ \hline
    $l$ & length of link \\ \hline
    $t$ & set of targets \\ \hline
    $o$ & set of obstacles \\ \hline
    $h$ & home base pose \\ \hline
    $\epsilon$ & index of the closest node to target's orientation segment \\ \hline
    $\theta^{(x)}_{\epsilon}$ & joint rotation around x-axis at the $\epsilon$-link toward target\\ \hline
    $\theta^{(y)}_{\epsilon}$ & joint rotation around y-axis at the $\epsilon$-link toward target\\ \hline
    $\bar{n}$ & number of active/folded links \\ \hline
    $l_{\bar{n}}$ & length of the last active/folded link \\ \hline
    $f_{1-2}$ & objective function: inverse kinematics \\ \hline
    $f_{3.1a}$ & objective function: number of links to segment \\ \hline
    $f_{3.1b}$ & objective function: number of links on segment \\ \hline
    $f_{3.2}$ & objective function: overall robot length \\ \hline
    $f_{3.3}$ & objective function: undulation \\ \hline
    \end{tabular}
}
\end{table}

The \textit{optimization problem} is defined with the terminology summarized in Table~\ref{tab:nomenclature}.

\subsection{Problem Formulation}
\label{sec:problem_formulation}

We begin the formulation with the following:

\begin{itemize}
    \item $n \in \mathbb{N}^+$, maximum link number, corresponding to the number of links the user is prepared to produce;
    \item $\theta^{(x)}$ and $\theta^{(y)} \in \mathbb{R}$, rotation of joint around x and y-axes, respectively, and constrained within a domain $\Delta_\theta $, which is determined by how much the robot can bend without breaking or buckling -- note that we decided to exclude the joint corresponding to the home base (i.e., the first) due to hardware observations so that the robot will grow from the base in the direction of its orientation and steer only from the second joint on; and 
    \item $l \in \mathbb{R}$, length of a link, constrained within a domain $\Delta_l$, and similar to the maximum link number $n$, it is influenced by production limitations (i.e., the overall size of the material sheet used for the body, the size of the cutting tool, etc.). Additionally, they have to be longer than the gripping mechanism to ensure stability on the end effector -- i.e., imagine the gripper that moves along the body as the robot everts, pushed by the end effector~\cite{jeong2020tip}; this mechanism has a specific length, which should be smaller than any link length because if a joint happens to be inside the gripper, that joint cannot bend.
\end{itemize}
Any start point of a link is defined as a \textit{node}, which also corresponds to a set of two rotational joints (around x and y) and a linear joint (for growth and retraction). Only the last endpoint (i.e., the end effector) is not considered a joint set but simply the robot's tip -- i.e., not a node. A design will have, at most, $n$ nodes plus the end effector.

The problem formulation also includes the \textit{task}, which is defined as:
\begin{itemize}
    \item $t$, target set, defined as an array of poses (i.e., for each target, three coordinates for position and two angles for desired reaching orientation);
    \item $o$, obstacle set, defined as an array of poses and geometrical properties\footnote{In this work, obstacles are modeled as cylinders oriented orthogonally with the floor, described by a radius, height, and the coordinates of its origin as the center of the base -- see in Section~\ref{sec:constraints}.}; and
    \item $h$, home base pose (three coordinates for position and two angles for orientation) -- although by default, we always place the home base to the origin of the reference frame, oriented toward the positive direction of the z-axis.
\end{itemize}

The output of the optimization consists of the following terms:
%problem consists of a design $\delta$, a solution $\psi$, and a set of $t$ configurations $\zeta_i$ (one for each target $i=1...|t|$):

\begin{itemize}
    \item configurations $\zeta_i$ (one for each target $i=1...|t|$), containing a column vector for the steering angles over the x-axis, y-axis, and the length of each link as in Eqn.~(\ref{eq:configuration});
    
    \begin{equation}\label{eq:configuration}
        \begin{aligned}
            \zeta = 
            \begin{bmatrix}
                \theta^{(x)}_1 & \theta^{(y)}_1 & l_1\\
                \vdots & \vdots & \vdots\\
                \theta^{(x)}_n & \theta^{(y)}_n & l_n\\
            \end{bmatrix}
        \end{aligned}
    \end{equation}
    
    \item a solution $\psi$ contains $|t|$ configurations, which provides the values of the decision variables and a description of the individuals' genotype, defined as in Eqn.~(\ref{eq:solution}) -- note that each pair of rows represent steering angles on x and y, respectively, with the last row being reserved for shared link lengths (resulting in a $(2 \cdot |t|+1) \times n$ matrix); and
    
    \begin{equation}\label{eq:solution}
        \begin{aligned}
            \psi = 
            \begin{bmatrix}
                \theta^{(x)}_{1,1} & \vspace{0.1cm} \multirow{2}{*}{$\cdots$} & \theta^{(x)}_{1,n}\\ 
                \theta^{(y)}_{1,1} & \vspace{0.1cm} & \theta^{(y)}_{1,n}\\
                \vdots & \vspace{0.1cm} \ddots & \vdots\\
                \theta^{(x)}_{|t|,1} & \vspace{0.1cm} \multirow{2}{*}{$\cdots$} & \theta^{(x)}_{|t|,n}\\
                \theta^{(y)}_{|t|,1} & \vspace{0.1cm} & \theta^{(y)}_{|t|,n}\\
                l_1 & \vspace{0.1cm} \cdots & l_n
            \end{bmatrix}
        \end{aligned}
    \end{equation}
        
    \item the final design $\delta$ defined as in Eqn.~(\ref{eq:design}), containing link lengths as they appear in $\psi$ -- i.e., one design for multiple configurations in the solution.

    \begin{equation}\label{eq:design}
        \begin{aligned}
            \delta = 
            \begin{bmatrix}
                l_1 & \cdots & l_n
            \end{bmatrix}
        \end{aligned}
    \end{equation}
    
\end{itemize}

% \subsection{Search Space}
% \label{sec:search_space}

% A higher number of links enhances the robot's dexterity and expands its workspace. Consequently, the optimization problem must handle a large search space $\omega$, which grows linearly with the number of targets 
% $|t|$, as it needs to optimize a single set of links for all targets.
% The search space $\omega$ is defined as in Eqn.~(\ref{eq:search_space}) when the order of targets is predetermined or if the robot returns to its base after reaching each target (i.e., the order of targets is irrelevant).

% \begin{equation}\label{eq:search_space}
%     \begin{aligned}
%         \omega = n^{\Delta_l} + 2 \cdot |t| \cdot n^{\Delta_\theta}
%     \end{aligned}
% \end{equation}

% When the order of targets is unknown, and the solution must also minimize the cost of transitioning between configurations, then the search space grows factorially, as every possible permutation of target sequences must be considered. In this work, we do not address the target order and focus solely on solving the basic optimization problem. The order can be considered after optimization with a proper motion planner like the one proposed by Altagiuri et al.~\cite{altagiuri2024motion} to determine the most efficient target-reaching sequence.

\subsection{Fitness Evaluation and Dynamic Genotype Change}
\label{sec:fitness}

We included the following objective functions:

\begin{itemize}
    \item $f_{1}$: For each configuration, minimize the distance between the end effector and the target -- inverse kinematic solver.
    \item $f_{2}$: For each configuration, minimize the displacement between the desired orientation segment and the robot's last link -- inverse kinematic solver.

    This is done by transforming configurations $\zeta$ in Eqn.~(\ref{eq:configuration}) from polar coordinates ($\zeta^p$, i.e., in joint space) to Cartesian coordinates ($\zeta^c$, i.e., in task space) through forward kinematics as in Eqn.~(\ref{eq:forwardkinematics}) -- details are reported in Section 1.1 of the Supplementary Material. The configuration $\zeta^c$ includes the end effector, resulting in $n+1$ nodes. 

    \begin{equation}\label{eq:forwardkinematics}
        \begin{aligned}
            \zeta^p = 
            \begin{bmatrix}
                \theta^{(x)}_1 & \theta^{(y)}_1 & l_1\\
                \vdots & \vdots & \vdots\\
                \theta^{(x)}_n & \theta^{(y)}_n & l_n\\
            \end{bmatrix}
            \quad
            %\underrightarrow{\emph{  forward kinematics  }}
            \xrightarrow[kinematics]{forward}
            \quad
            \zeta^c = 
            \begin{bmatrix}
                x_1 & y_1 & z_1\\
                \vdots & \vdots & \vdots\\
                x_{n+1} & y_{n+1} & z_{n+1} 
            \end{bmatrix}
        \end{aligned}
    \end{equation}
    
    \begin{itemize}
        \item $f_{1-2}$: We merged these two objectives into a single one, which is to minimize the distance between the closest node to the target's orientation segment and the segment itself, allowing the robot to correctly aim for the target's pose as shown in Fig.~\ref{fig:min_nodetargetorient}. This is calculated according to Eqn.~(\ref{eq:f1f2}) -- the distance from a node to the $i$-th target's orientation segment $s_i$ is reported in Algorithm S.3 of the Supplementary Material.

        \begin{equation}\label{eq:f1f2}
            \begin{aligned}
                f_{1-2} = \sum_{i=1}^{|t|} {\min_{\forall j \in n}{(\texttt{pointSegmentDistance}(\zeta^c_{i,j+1},s_i)})}
            \end{aligned}
        \end{equation}

        This formulation requires some additional information to the basic genotype described in Eqn.~(\ref{eq:solution}): once the closest node to the target's orientation segment is found (at index $\epsilon$), the remainder of the configuration's link is unused (the dotted links in Fig.~\ref{fig:min_nodetargetorient}), and we replace them with links that go straight to the target. This results in a different orientation from the angles specified for the $\epsilon$-th joint (defined as $\theta^{(x)}_{\epsilon}$ and $\theta^{(y)}_{\epsilon}$), a number of \textit{actually used} links in the overall configuration ($\bar{n}$), and the length of the last link that does not necessarily need to be fully unfolded ($l_{\bar{n}}$). The procedures for retrieving these values are reported in the Supplementary Material. 
    \end{itemize}

    \begin{figure*}[h!]
        \centering
        \subfigure[\protect\url{}\label{fig:min_nodetargetorient}Inverse Kinematic Solver]%
        {\includegraphics[height=6.5cm]{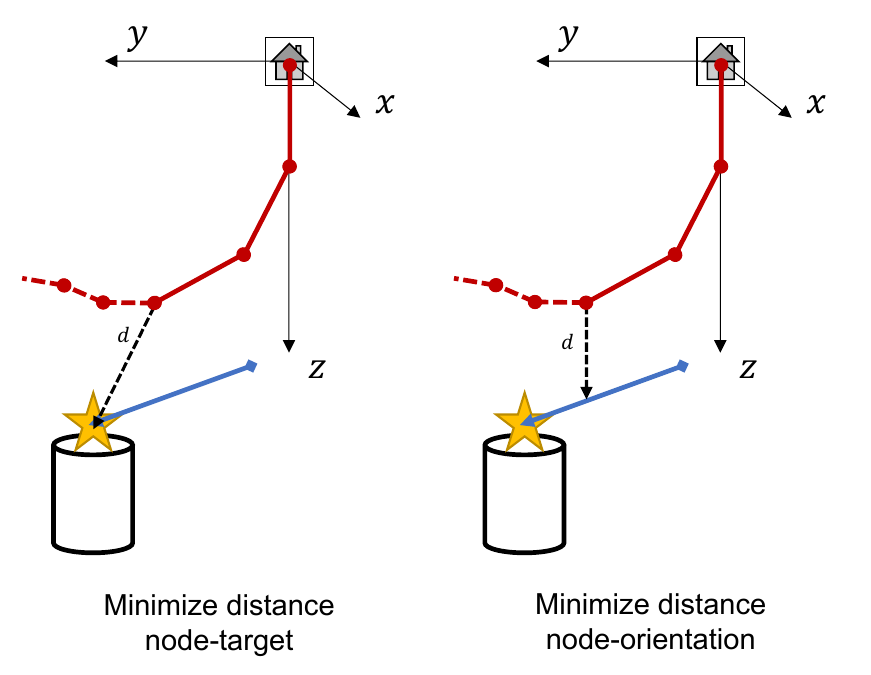}}
        \subfigure[\protect\url{}\label{fig:phenotype}Phenotype Description]%
        {\includegraphics[height=6.5cm]{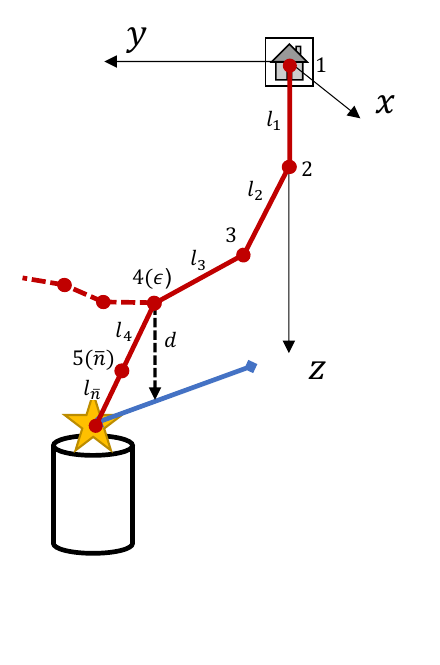}}
        \caption{(a) When the inverse kinematic solver minimizes the distance node-target $d$, there is no guarantee for the robot to have the last link aligned with the desired target orientation; however, when the solver minimizes the distance node-orientation, the robot will reach the desired orientation and can simply grow straight to the target after that. (b) The phenotype codified by a solution $\psi$ ignores every link after $\epsilon$ originally generated in the genotype and aims directly to the target for the remaining $\bar{n}-\epsilon$ links. The last link might not be fully everted, with a length smaller than the one originally generated in the genotype -- indicated with the term $l_{\bar{n}}$.}
        \label{fig:inverse_kinematics}
    \end{figure*}

    \item $f_3$: For each configuration, minimize the robot's components to promote a quicker and more cost-effective manufacture. We split this objective into the following:

        \begin{itemize}
            \item $f_{3.1a}$: For each configuration, minimize the number of links needed to reach the target's orientation segment, as in Eqn.~(\ref{eq:f_31a}).
            
            \begin{equation}\label{eq:f_31a}
                \begin{aligned}
                    f_{3.1a} = \sum_{i=1}^{|t|} (\epsilon_i - 1)
                \end{aligned}
            \end{equation}
            
            \item $f_{3.1b}$: For each configuration, minimize the number of links lying on the target's orientation segment, as in Eqn.~(\ref{eq:f_31b}).

            \begin{equation}\label{eq:f_31b}
                \begin{aligned}
                    f_{3.1b} = \sum_{i=1}^{|t|} ( \bar{n}_i - \left(\epsilon_i - 1 \right))
                \end{aligned}
            \end{equation}
            
            \item $f_{3.2}$: Minimize the overall length of the robot, which is the longest robot's length among all configurations, as in Eqn.~(\ref{eq:f_32}).

            \begin{equation}\label{eq:f_32}
                \begin{aligned}
                    f_{3.2} = {\max_{\forall i \in |t|}{\left( \left( \sum_{j=1}^{\bar{n}_i - 1}{l_j} \right) + l_{\bar{n}_i} \right)}}
                \end{aligned}
            \end{equation}

            \begin{figure*}[t!]
                \centering
                \subfigure[\protect\url{}\label{fig:fit_wiggly}Wavy robot]%
                {\includegraphics[height=6cm]{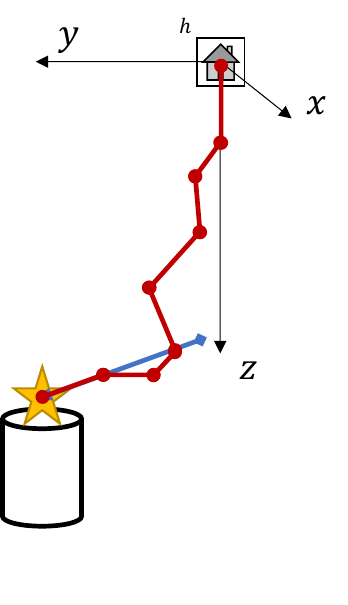}}
                \subfigure[\protect\url{}\label{fig:tradeoff_nodesOrient}Trade-off nodes/orientation]%
                {\includegraphics[height=6cm]{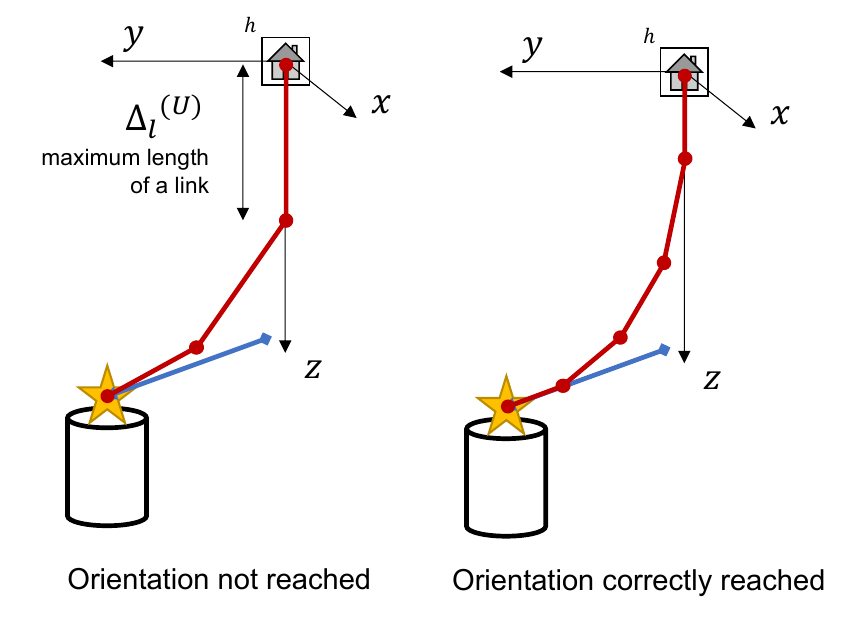}}
                \caption{(a) A poorly designed robot featuring a wavy configuration. (b) Reducing the number of links and increasing their length helps solve this issue, a trade-off is necessary to make sure the robot smoothly reaches the target.}
                \label{fig:tradeoffs}
            \end{figure*}
            
            \item $f_{3.3}$: For each configuration, minimize the directional change of the links, or \textit{undulation}, to avoid ``wavy'' configurations are shown in Fig.~\ref{fig:tradeoffs}. This is evaluated in percentage over the number of links, as shown in Eqn.~(\ref{eq:f_33}) -- note that the formula checks the change in the angle between the $j$-th and $(j+1)$-th link in both the $xz$ and $yz$-plane projections ($\Delta \theta^{(xz)}$ and $\Delta \theta^{(yz)}$, respectively). 

            \begin{equation}\label{eq:f_33}
                \begin{aligned}
                    f_{3.3} = 
                    \frac{\sum_{i=1}^{|t|}{\left( \frac{\forall k \in [xz,yz] , \;  \sum_{j=1}^{\epsilon_i - 1}{ \left[ \sign{(\theta^{(k)}_{i,j})} \neq \sign{(\theta^{(k)}_{i,j+1})} \right]}}{2 \cdot \epsilon_i} \right)}}{|t|} \cdot 100
                \end{aligned}
            \end{equation}

        \end{itemize}
\end{itemize}

Since this evaluation dynamically adds extra information to the genotype, it is necessary to extend the solution matrix $\psi$ by including, for each configuration:

\begin{itemize}
    \item $\epsilon$, the index of the configuration node/joint that is closest to the target's orientation segment;
    \item $\theta^{(x)}_{\epsilon}$ and $\theta^{(y)}_{\epsilon}$, the angles on the x and y-axes defining the orientation for the $\epsilon$-th link, dynamically set to go in the direction of the target;
    \item $\bar{n}$, the number of active nodes/links (with $\bar{n} \leq n$) -- namely the index of the \textit{last active node} of a configuration; and
    \item $l_{\bar{n}}$, the last link's length.
\end{itemize}
The complete matrix $\psi$ is defined as in Eqn.~(\ref{eq:solution_withExtra}), and the respective phenotype that it represents is shown in Fig.~\ref{fig:phenotype} -- example reports one configuration only for the sake of clarity. Note that, being a matrix, some elements are set to zero as they are unused: (i) the second row of each configuration, which defines rotations on the y-axis, only contains the last angle rotation $\theta^{(y)}_{\epsilon}$; and (ii) the last row, which defines the lengths of the links, is fixed for a design and do not need any extra information.

\begin{equation}\label{eq:solution_withExtra}
    \begin{aligned}
        \psi = 
        \begin{bmatrix} 
            \begin{bmatrix} 
                \theta^{(x)}_{1,1} & \vspace{0.1cm} \multirow{2}{*}{$\cdots$} & \theta^{(x)}_{1,n}\\ 
                \theta^{(y)}_{1,1} & \vspace{0.1cm} & \theta^{(y)}_{1,n}\\
                \vdots & \vspace{0.1cm} \ddots & \vdots\\
                \theta^{(x)}_{|t|,1} & \vspace{0.1cm} \multirow{2}{*}{$\cdots$} & \theta^{(x)}_{|t|,n}\\
                \theta^{(y)}_{|t|,1} & \vspace{0.1cm} & \theta^{(y)}_{|t|,n}\\
                l_1 & \vspace{0.1cm} \cdots & l_n
            \end{bmatrix}
            &
            \begin{bmatrix}
                \epsilon_1 & \vspace{0.1cm} \theta^{(x)}_{\epsilon,1} & \bar{n}_1 & l_{\bar{n},1}\\
                0 & \vspace{0.1cm} \theta^{(y)}_{\epsilon,1} & 0 & 0\\
                \vdots & \vspace{0.1cm} \vdots & \vdots & \vdots \\
                \epsilon_{|t|} & \vspace{0.1cm} \theta^{(x)}_{\epsilon,|t|} & \bar{n}_t & l_{\bar{n},|t|}\\
                0 & \vspace{0.1cm} \theta^{(y)}_{\epsilon,|t|} & 0 & 0\\
                0 & \vspace{0.1cm} 0 & 0 & 0
            \end{bmatrix}
        \end{bmatrix}
    \end{aligned}
\end{equation}

\begin{figure}[b!]
    \centering
    {\includegraphics[width=7cm]{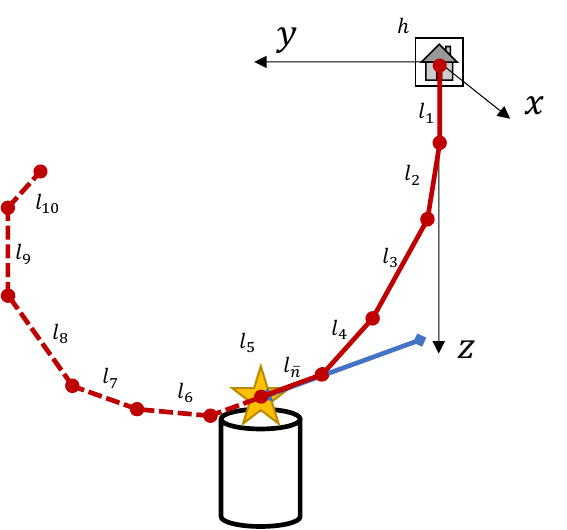}}
    \caption{The genotype described in Eqn.~(\ref{eq:solution}) yields the phenotype in the example, with a budget of $n=10$ links. However, only five of them are required to correctly reach the target ($\bar{n} = 5$), an information included in the genotype of Eqn.~(\ref{eq:solution_withExtra}).}
    \label{fig:everted_configuration}
\end{figure}

\subsection{Constraints}
\label{sec:constraints}

\begin{figure}[t!]
    \centering
    \subfigure[\protect\url{}\label{fig:constraint_1_4}]%
    {\includegraphics[width=4cm]{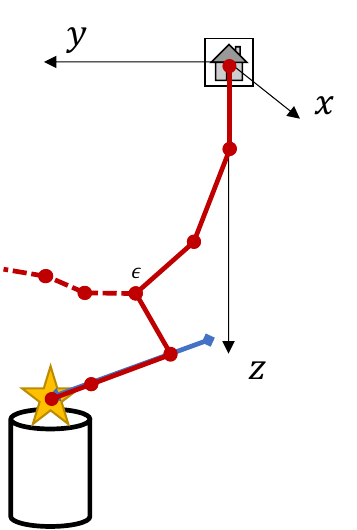}}
    \subfigure[\protect\url{}\label{fig:constraint_5}]%
    {\includegraphics[width=4cm]{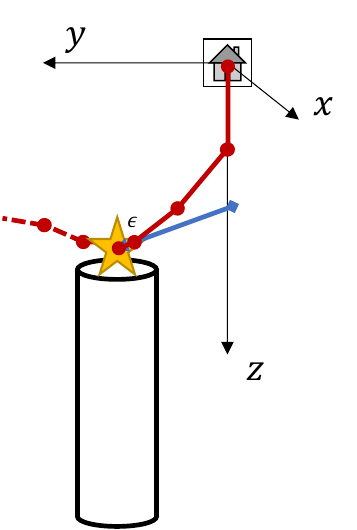}}
    \subfigure[\protect\url{}\label{fig:constraint_6_7}]%
    {\includegraphics[width=4cm]{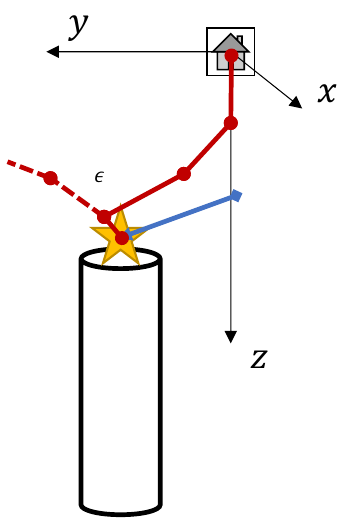}}\\
    \subfigure[\protect\url{}\label{fig:constraint_7}]%
    {\includegraphics[width=4cm]{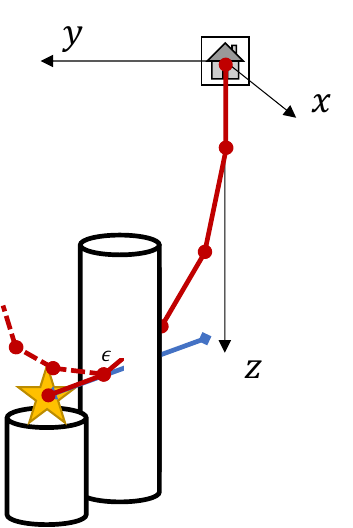}}
    \subfigure[\protect\url{}\label{fig:constraint_9}]%
    {\includegraphics[width=4cm]{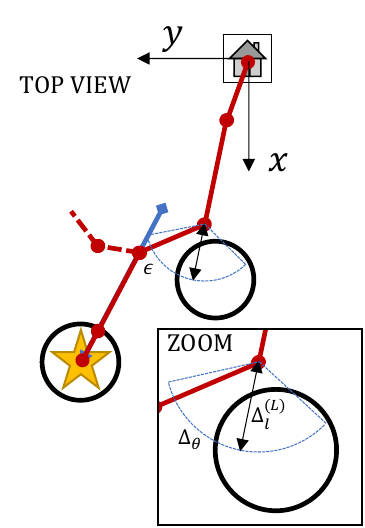}}
    \caption{Examples of solutions violating constraints -- i.e., infeasible. (a) Excessive steering when reaching the target; (b) a small final link that makes it impossible for the robot to carry a gripping mechanism; (c) overshooting the target, resulting in excessive angles; (d) an obstacle collision; and (e) a collision non detected in the configuration but as result of motion, which requires a link to be everted before steering. }
    \label{fig:constraints}
\end{figure}

The problem's constraints are similar to the 2D case presented in the previous work~\cite{stroppa2024design}, with the respective additions of one more dimension for the rotational angles and one more constraint on obstacle collision that was not previously -- and erroneously -- considered.
Every configuration in the solutions is subject to eight kinematic constraints (depicted in Fig.~\ref{fig:constraints}) in addition to the decision-variable bounds described in Section~\ref{sec:problem_formulation}.

The first four constraints concern the angles at the $\epsilon$-th node, which are set by the evaluation function such that, from that point on, the robot grows in the direction of the target -- as described in Section~\ref{sec:fitness}. Since these angles are geometrically calculated, there is no guarantee that they respect the bounds $\Delta_{\theta}$, as illustrated in Fig.~\ref{fig:constraint_1_4}; therefore, we included the four inequality constraints defined in Eqn.~(\ref{eq:constraint_1_4}). Note that the bound $\Delta_{\theta}$ can also differ between axes, but for the sake of simplicity, we will report it as the same value.

\begin{equation}\label{eq:constraint_1_4}
    \begin{aligned}
        \theta^{(x)}_{\epsilon} \geq {\Delta_\theta}^{(L)} \wedge \theta^{(x)}_{\epsilon} \leq {\Delta_\theta}^{(U)} \\
        \theta^{(y)}_{\epsilon} \geq {\Delta_\theta}^{(L)} \wedge \theta^{(y)}_{\epsilon} \leq {\Delta_\theta}^{(U)}
    \end{aligned}
\end{equation}

The fifth constraint considers the case in which the length of the last link is smaller than the lower bound ${\Delta_l}^{(L)}$, as shown in Fig.~\ref{fig:constraint_5}. This is an issue only when the final link needs to adjust its direction to align with the target's orientation segment -- i.e., there is only one link in on the segment, or $f_{3.1b} = 1$, Eqn.~(\ref{eq:f_31b}): these soft growing robots are likely equipped with a gripper device that moves with the end effector along the robot's body (e.g., the tip mount proposed by Jeong et al.~\cite{jeong2020tip} or the mechanism to prevent buckling proposed by Coad et al.~\cite{coad2020retraction}). For this reason, it is essential to account for the necessary space to position this gripper properly on the target's orientation segment for accurate grasping. The respective inequality constraint is expressed in Eqn.~(\ref{eq:constraint_5}).

\begin{equation}\label{eq:constraint_5}
    \begin{aligned}
        \text{if} \; f_{3.1b} = 1 \text{,} \quad  l_{\bar{n}} \geq {\Delta_l}^{(L)}
        %\left[ f_{3.1b} = 1 \right], \quad  l_{\bar{n}} \geq {\Delta_l}^{(L)}
    \end{aligned}
\end{equation}

The sixth constraint handles an edge case of the point-segment distance method of Eqn.~(\ref{eq:f1f2}) -- details of the procedure are reported in Section 1.2 of the Supplementary Material. Since this process may designate one of the segment's endpoint as the nearest point to $\epsilon$ -- rather than $\epsilon$'s projection on the segment -- if that endpoint corresponds to the target, the resulting solution would resemble the configuration shown in Fig.~\ref{fig:constraint_6_7}.
Specifically, we refer to the case at line 19 of Algorithm S.3 presented in the Supplementary Material. 
To avoid this issue, we established an additional inequality constraint as defined in Eqn.~(\ref{eq:constraint_6}). This constraint checks whether the angle width between the end of each configuration $\zeta_i$ and the target's orientation segment $s_i$ achieves or exceeds $\pi/2$. Specifically, the target's orientation segment $s_i$ (mentioned in Eqn.~(\ref{eq:f1f2}) as well) has two endpoints: $s_{i,1}$ is the coordinates of its respective target $t_{i,1\rightarrow3}$, and  $s_{i,2}$ is any point on the straight line where $s_i$ lies. The angle between these two segments must be smaller than $\pi/2$ for the configuration to be feasible; otherwise, it means that the robot overshot and is trying to align with the target rather than its orientation segment.

\begin{equation}\label{eq:constraint_6}
    \begin{aligned}
        \cos^{-1} \left( \frac{(\zeta^{c}_{\epsilon_i} - s_{i,1}) \cdot (s_{i,2} - s_{i,1})}{\|\zeta^{c}_{\epsilon_i} - s_{i,1}\| \cdot \|s_{i,2} - s_{i,1}\|} \right) <  \frac{\pi}{2}, \text{ where } i \in [ 1,..., |t| ]
\end{aligned}
\end{equation}

Lastly, we impose two constraints related to obstacle collisions. The first one imposes that no collision is generated in the solution, as a robot colliding with an obstacle would be an infeasible solution -- as illustrated in Fig.~\ref{fig:constraint_7}. The equality constraint in Eqn.~(\ref{eq:constraint_8}) evaluates the intersections between the links in $\psi$ and the obstacles in the set $o$. The full procedure is reported in Section 1.4 of the Supplementary Material, and it operates with a complexity of $\mathcal{O}(\overline{n} \cdot |o| \cdot |t|)$. To simplify the process and reduce computational overhead, we model every obstacle as a cylinder due to the ease of detecting intersections between a segment and a cylinder during a collision~\cite{dong2015obstacle}. More complex obstacles can be modeled by combining multiple cylinders: a pillar is represented by a single cylinder; a wall by a line of cylinders; and a hole in a wall by stacking cylinders in a bottom layer, a top layer, and two taller cylinders at the sides, forming a hole in the middle. During problem setup, users can select a larger radius than that of the obstacle's circumscribed cylinder to prevent contact between the obstacle and the manipulator.
Note that avoiding obstacles might also be considered as another objective function: one can (i) minimize obstacle collisions or (ii) maximize the distance between any robot link and obstacles. However, the first option would consider a solution with obstacle collision as feasible, which deceives our purposes as we do not want collisions to happen at all. On the other hand, the second option might not work if a target is surrounded by obstacles (e.g., the target is in an open box). Therefore, considering obstacle collisions as a constraint is the best option for our optimization problem.

\begin{equation}\label{eq:constraint_8}
    \begin{aligned}
        \texttt{intersections}(\psi,o) = 0
    \end{aligned}
\end{equation}

The last constraint regards a possible obstacle collision when actuating the manipulator -- i.e., the static configuration $\zeta_i$ does not collide with obstacles, but the actuation to move to $\zeta_i$ from any other configuration $\zeta_j$ will cause a collision. This is because of how the manipulator is actuated: a link must grow of at least a minimum length (i.e., the lower bound $\Delta^{(L)}_l$) before is enabled to steer. Similarly to the constraint expressed in Eqn.~(\ref{eq:constraint_5}) and Fig.~\ref{fig:constraint_5}, this prevents the robot from steering while the joint is within the tip mount, avoiding possible damage. Fig.~\ref{fig:constraint_9} shows an example: if the configuration requires the robot to grow of a length equal to $\Delta^{(L)}_l$ within the obstacle, then we have a collision -- note that the figure represents a projection in 2D of a spherical cone. A motion planner could actually find a feasible alternative path without colliding with the obstacle (i.e., by bending one of the previous joints); however, checking this during optimization would require running the motion planner~\cite{altagiuri2024motion}, which would require too much in terms of computational resources. Therefore, we simply mark a configuration like this one as infeasible. This constraint is summarized in Eqn.~(\ref{eq:constraint_9}), which imposes the number of collisions while steering to be equal to zero. More details are provided in Section 1.5 of the Supplementary Material.

\begin{equation}\label{eq:constraint_9}
    \begin{aligned}
        \texttt{cannotSteer}(\psi,o) = 0
    \end{aligned}
\end{equation}

\section{Experiments}
\label{sec:experiments}

We ran a set of experiments to (i) assess whether the optimizer can retrieve optimal solutions on tasks with large dimensionality and many obstacles; (ii) determine which EC algorithm is the best fit for solving our design problem among four different options; and (iii) evaluate the robustness of the Rank Partitioning algorithm when included in the loop of different EC algorithms.  

\subsection{Optimization Methods}
\label{sec:optimization_methods}

To solve this design optimization problem, we employ the \textit{Rank Partitioning} method, which was introduced in the optimizer for the 2D case~\cite{stroppa2024design}. Rank partitioning is a non-parametric survival strategy for multi-objective EC algorithms (or generically for population-based optimization methods) that ranks each individual based on a predefined order of the objectives. It sorts the population for the first objective (i.e., the one with the highest priority) and then creates partitions based on their objective rank; then, it sorts each partition for the second objective, and so on. Finally, it returns only one solution: the optimal based on the given order of priorities -- specifically, the priority order we selected for our objectives is $f_{1-2}$ (to ensure the robot reaches the targets), $f_{3.1a}$ (to avoid configurations roaming before reaching the correct orientation), $f_{3.2}$ (to prevent wavy configurations), $f_{3.1b}$ and $f_{3.2}$ (to minimize material waste). 
The advantages of such a method are: (i) it allows us to retrieve the optimal solution based on our preference without compromising trade-offs; (ii) there is no need to retrieve the full Pareto optimal front, saving computational resources; and (iii) it outperforms weighted sum based methods while without featuring their disadvantages~\cite{miettinen2012nonlinear,hwang2012multiple}. In this section, we seek evidence for a fourth advantage: this method can be applied to \textit{any} EC algorithm, or any multi or many-objective optimization (i.e., with more than three objectives~\cite{deb2013evolutionary}).
More details about Rank Partitioning are reported in Section 2 of the Supplementary Material or in our previous work~\cite{stroppa2024design}.

\IncMargin{1em}
\begin{algorithm}[t!]
	\SetKwFunction{Init}{randomInitialization}
        \SetKwFunction{Eval}{evaluation}
        \SetKwFunction{Const}{constraintChecking}
        \SetKwFunction{Rank}{rankPartitioning}
        \SetKwFunction{Sel}{selection}
        \SetKwFunction{Var}{variation}
        \SetKwFunction{Sur}{survival}
        \SetKwFunction{Update}{updateFittest}
	\SetKwInOut{Input}{input}
	\SetKwInOut{Output}{output}
	\Input{The optimization problem $op$ (in our case, described by $t$, $h$, $o$, $n$, $\Delta_{\theta}$, $\Delta_{l}$) including a set of constraints $c$}
        \Input{The EA parameter such as number of individuals $N$ and number of generations $G$}
	\Output{The most fitting individual $best$ and its fitness $f^{best}$}
	\BlankLine		
	
	\Begin{	

    	$pop \gets \Init(N)$\\
            $F^{pop} \gets \Eval(op,c,pop)$\\
            %$F^{pop} \gets \Const(c,pop,F^{pop})$\\
            $F^{pop} \gets \Rank(F^{pop})$\\
            \For{$i \gets 1 $ \KwTo $ G$}
            {
                %$mat \gets \Sel(pop)$\\
                $off \gets \Var(pop, F^{pop})$\\
                $F^{off} \gets \Eval(op,c,off)$\\
                %$F^{off} \gets \Const(c,off,F^{off})$\\

                \uIf{apply elitism}
                {
                    $F^{pop} \cup F^{off} \gets \Rank(F^{pop} \cup F^{off})$\\
                    $pop, F^{pop} \gets \Sur(pop,off,F^{pop} \cup F^{off})$\\
                }
                \Else
                {
                    $F^{off} \gets \Rank(F^{off})$\\
                    $pop \gets off$\\
                    $F^{pop} \gets F^{off}$\\
                }
                %$best, f^{best}  \gets \Update(pop,F^{pop},best)$\\
                $best, f^{best}  \gets \Rank(f^{best} \cup F^{pop}_1)$\\
            }
		\KwRet{$best$, $f^{best}$}\\
	}
	
	\caption{EC framework with Rank Partitioning}\label{alg:ea_framework}
\end{algorithm}\DecMargin{1em}

We tested the design optimizer on four different EC algorithms: Genetic Algorithm (GA)~\cite{goldberg1990real}, Particle Swarm Optimization (PSO)~\cite{kennedy1995particle}, Differential Evolution (DE)~\cite{storn1997differential}, and Big Bang-Big Crunch algorithm (BBBC)~\cite{erol2006new}. 
Each of these EC algorithm has its own parameters, which are described in the following subsections; however, some of the parameters are shared as they all comply with the basic framework of EC algorithms, reported in Algorithm~\ref{alg:ea_framework}: an initial population $pop$ is initialized with $N$ random individuals (line 2), evaluated returning the fitness of each individual $F^{pop}$ penalized in case of constraint violation (line 3), ranked based on the Rank Partitioning algorithm (line 4, runs Algorithm S.11 of the Supplementary Material), and then it undergoes the generational loop for $G$ iterations (line 5-14) in which variation operators generate a new population of offspring $off$ (line 6), evaluate them returning their fitness $F^{off}$ with penalty in case of constraint violation (line 7), and select which individuals among $pop$ and $off$ will survive and be part of the next generation based on the survival scheme (lines 8-14) -- specifically, when elitism is applied, the Rank Partitioning algorithm is executed on both populations (line 9), otherwise only on the offspring (line 12). Note that, with Rank Partitioning, the best individual to be saved and returned by the method is not the individual with minimum/maximum fitness value, but it is the one having rank $=1$ after executing Rank Partitioning between the previous best and the best of the current generation (line 15) -- regardless on whether the method is elitist or not.
According to preliminary evaluations\footnote{For better readability, we do not report any explicit detail of this preliminary study except for the p-values. However, the data can be provided at readers' request.}, we set only GA to be elitist (with elitism being significantly better than non-elitist, $p<0.01$), whereas the other EC algorithms were all set to be non-elitist -- PSO and DE were found to be significantly better when elitism is not applied ($p<0.01$ and $p=0.0179$, respectively), and no significant difference was observed with BBBC ($p=0.3793$). 
%Note that, in our specific case, the Rank Partitioning algorithm is executed after each evaluation and constraint checking on a single population and re-executed during the survival stage to rank individuals from both populations.
Table~\ref{tab:param_shared} reports the parameters and settings shared among all EC algorithms, with the values we used in the study.

\begin{table}[h!]
\centering
\caption{Experimental conditions and parameters (shared among EC algorithms)}
\label{tab:param_shared}
%\vspace{0.2cm}
    \begin{tabular}{lc}
    \hline
    \multicolumn{1}{c}{\textbf{Parameter}} & \textbf{Value} \\ \hline
    Population Size ($N$) & $500$ \\ \hline
    Max Number of Generations ($G$) & $500$ \\ \hline
    %Constraint Handling & \makecell{Adaptive Penalty~\cite{coello2002theoretical,ben1997genetic} \\ $\lambda(g=1) = \todo{1.0}$, $p = 10$ \\ $\beta_1 = 1.05$, $\beta_2 = 1.75$} \\  \hline
    Constraint Handling & \makecell{Static Penalty~\cite{homaifar1994constrained,coello2002theoretical} \\ $R=100$} \\  \hline
    Rank Partitioning Bin Size & \makecell{for $f_{1-2} = 0.5$ \\ for $f_{3.2} = 5.0$} \\ \hline
    \end{tabular}
\end{table}

The population size $N$ was chosen based on the results of the previous 2D work~\cite{stroppa2024design}, whereas the maximum number of generations $G$ was set to a high number to ensure the convergence of the algorithms -- i.e., the value of the best solution retrieved by the optimizer stopped changing before the run terminated.
%The population size $N$ was chosen based on the results of the previous 2D work~\cite{stroppa2024design}, whereas the maximum number of generations $G$ was chosen by observing the convergence of the algorithms (we tested it on a GA): we ran a medium-difficulty task (Task 2 described in Section~\ref{sec:tasks}) twenty times with $G=500$ and $N=500$, observed the change in variance taken over a span of ten generations for all objectives and recorded the generation number when the variance change goes below $1.0$. This provides an approximation of the generation where the algorithm converged, and we set the final value of $G$, adding a $20\%$ to this value.

For constraint handling, we used the Static Penalty method~\cite{homaifar1994constrained,coello2002theoretical}. With this method, infeasible solutions are penalized with a fixed penalty factor $R$. We empirically set the value of $R$ to $100$ for every constraint, a value that allows the optimizer to retrieve solutions without fluctuating between feasible and infeasible at each iteration.
%For constraint handling, we used the Adaptive Penalty method described in the survey by Coello Coello~\cite{coello2002theoretical}, also referred to as the Nonlinear Penalty method by the original authors Ben and Bean~\cite{ben1997genetic}. With this method, infeasible solutions are penalized with a penalty coefficient $\lambda(g)$ that changes based on the trend in the last $p$ generations: if the best individual was always infeasible, increase the penalty; if the best individual was always feasible, decrease the penalty; otherwise, keep the penalty constant -- the values of $\beta_1$ and $\beta_2$, which govern how much $\lambda(g)$ changes, were chosen empirically. During the pre-experimental phase, we observed a consistent improvement in the solutions with respect to the ones retrieved with static penalty~\cite{coello2002theoretical}, which was employed in the previous 2D work~\cite{stroppa2024design}.

Finally, we chose Rank Partitioning's bin sizes for the continuous-to-discrete objectives based on the results of the previous 2D work~\cite{stroppa2024design}. Specifically, for $f_{1-2}$ (inverse kinematics), a greater bin size corresponds to a significant decrease in the number of links; however, for some tasks where flexibility is required, a smaller bin size can be preferred. This allows the robot to maneuver comfortably between obstacles. For $f_{3.2}$ (total robot length), 
we observed the average total link lengths from the twenty pre-runs we used to estimate the convergence generation ($200 \pm 78$) and compared them to the ones in the 2D work ($216 \pm 63$). Since these robot lengths are similar, we selected the same bin size value.

\subsubsection{Genetic Algorithm (GA)}
\label{sec:ga}
GAs are designed for optimization in continuous or discrete domains by simulating the process of natural evolution~\cite{goldberg1990real}. In GAs, candidate solutions, known as individuals, are typically represented as encoded vectors or chromosomes. The algorithm iteratively evolves the population through genetic operations: selection, crossover, and mutation. During selection, individuals with higher fitness are chosen to propagate their traits to the next generation. Crossover combines pairs of selected individuals to create offspring, allowing for the exploration of new regions in the search space. Mutation introduces small, random changes to individual genes, helping to maintain diversity and avoid premature convergence. By iteratively applying these operations, GAs balance exploration and exploitation of the search space, guiding the population toward optimal or near-optimal solutions over successive generations. This flexibility makes GAs particularly effective for complex, multimodal optimization problems.

\begin{table}[h!]
\centering
\caption{Experimental conditions and parameters (GA)}
\label{tab:param_ga}
\resizebox{0.9\columnwidth}{!}{%
%\vspace{0.2cm}
    \begin{tabular}{lc}
    \hline
    \multicolumn{1}{c}{\textbf{Parameter}} & \textbf{Value} \\ \hline
    Selection Type & Binary Tournament~\cite{goldberg1991comparative} \\ \hline
    Crossover Type & \makecell{blx-$\alpha$~\cite{eshelman1993real} \\ $\alpha = 0.500, 0.419, 0.381$} \\ \hline
    Crossover Probability ($p_c$) & 1.0 \\  \hline
    Mutation Type & Random~\cite{michalewicz1992genetic} \\  \hline
    Mutation Probability ($p_m$) & 0.2, 0.4, 0.6, Dynamic $1.0 \rightarrow 0.0$ \\  \hline
    Survival Type & Elitist ($\mu + \lambda$ schema)~\cite{beyer2002evolution}\\  \hline
    %Survival Type & \makecell{Non Elitist,\\ Elitist ($\mu + \lambda$ schema)~\cite{beyer2002evolution}} \\  \hline
    \end{tabular}
}
\end{table}

Table~\ref{tab:param_ga} reports the parameter settings we used for GA. We used the binary tournament for selecting individuals undergoing variation~\cite{goldberg1991comparative}, the blx-$\alpha$ for crossover~\cite{eshelman1993real}, random mutation~\cite{michalewicz1992genetic}, and both elitist and non-elitist methods for survival. Specifically, we tested three values for $\alpha$ in the crossover, which dictates how far the offspring are generated from the parents -- these values were selected based on a comparative study on self-adaptation among crossover methods~\cite{beyer2000desired} to determine the range of probability distribution between parents. Since the effects of mutation cannot be predicted on the specific problem (i.e., what happens with different degrees of exploration), we tested four different probabilities of mutation, including a strategy that dynamically and linearly decreases it from 1 to 0 with the generation counter (i.e., full exploration at early generations, full exploitation toward the end). 
%The non-elitist survival method simply replaces the entire population with the offspring (i.e., stores the best solution without having it undergo the genetic operations), whereas the alternative 
The elitist survival method employs the $\mu + \lambda$ schema~\cite{beyer2002evolution}, which (i) merges the old population and the offspring, (ii) sorts them by fitness (in our case, with Rank Partitioning), and (iii) selects only the $N$ best individuals.

\subsubsection{Particle Swarm Optimization (PSO)}
\label{sec:pso}
PSO is designed for optimization in continuous domains by representing each candidate solution, or particle, as a position in a multi-dimensional space~\cite{kennedy1995particle}. Unlike other algorithms that rely on crossover or mutation, PSO uses the concept of velocity to iteratively update the positions of particles. Each particle adjusts its velocity based on both its own previous best position and the best-known positions of its neighbors or the entire swarm. This dynamic adjustment allows particles to explore the search space more effectively. The search directions in PSO are influenced by both local and global information, promoting a balance between exploration and exploitation. This process enables particles to converge toward optimal or near-optimal solutions while avoiding getting trapped in local optima, making PSO particularly effective for solving complex, nonlinear optimization problems.

\begin{table}[h!]
\centering
\caption{Experimental conditions and parameters (PSO)}
\label{tab:param_pso}
\resizebox{0.9\columnwidth}{!}{%
%\vspace{0.2cm}
    \begin{tabular}{lc}
    \hline
    \multicolumn{1}{c}{\textbf{Parameter}} & \textbf{Value} \\ \hline
    Inertia Weight ($\omega$) & $0.6, 0.8, 1.0$~\cite{eberhart2001tracking} \\ \hline
    Cognitive Constant ($c_1$) & $0.5, 2.5, 5.0$~\cite{zhou2011randomization,ratnaweera2004self} \\ \hline
    Social Constant ($c_2$) & $0.5, 2.5, 5.0$~\cite{zhou2011randomization,ratnaweera2004self} \\ \hline
    \end{tabular}
}
\end{table}

Table~\ref{tab:param_pso} reports the parameter settings we used for PSO. The inertia weight $\omega$ is between [0.5, 1.0] after the random $\omega$ factor formula~\cite{eberhart2001tracking}. As for $c_1$ and $c_2$,  Ratnaweera et al.~\cite{ratnaweera2004self} reported the best range for those constants being [0.5, 2.5], with 5.0 being the best parameter suggested by Zhou et al.~\cite{zhou2011randomization} %We used the binary tournament for selecting individuals undergoing variation~\cite{goldberg1991comparative}, the blx-$\alpha$ for crossover~\cite{eshelman1993real}, random mutation~\cite{michalewicz1992genetic}, and both elitist and non-elitist methods for survival. Specifically, we tested three values for $\alpha$ in the crossover, which dictates how far the offspring are generated from the parents -- these values were selected based on a comparative study on self-adaptation among crossover methods~\cite{beyer2000desired} to determine the range of probability distribution between parents. Since the effects of mutation cannot be predicted on the specific problem (i.e., what happens with different degrees of exploration), we tested four different probabilities of mutation, including a strategy that dynamically and linearly decreases it from 1 to 0 with the generation counter (i.e., full exploration at early generations, full exploitation toward the end). The non-elitist survival method simply replaced the entire population with the offspring (i.e., stores the best solution without having it undergo the genetic operations), whereas the alternative elitist method employs the $\mu + \lambda$ schema~\cite{beyer2002evolution} (i.e., merges the old population and the offspring, sorts them by fitness with Rank Partitioning, and selects only the $N$ best individuals).

\subsubsection{Differential Evolution (DE)} 
\label{sec:de}
DE is specifically designed for optimization in continuous domains by representing each potential solution, or individual, as a multi-dimensional vector~\cite{storn1997differential}. Perturbations are introduced into the population through vector differences, allowing DE to explore the search space efficiently. A key distinction from GAs lies in its mutation strategy: instead of relying on predefined probability distributions or crossover rates, DE generates mutations based on the differences between randomly selected vectors from the current population. This dynamic and adaptive approach allows the search directions to be guided by the relative positions of individuals, promoting diverse exploration and reducing the chances of premature convergence to local optima. Additionally, DE’s mechanism of mutation and recombination maintains the balance between exploration and exploitation, which is crucial for solving complex, nonlinear optimization problems.

\begin{table}[h!]
\centering
\caption{Experimental conditions and parameters (DE)}
\label{tab:param_de}
\resizebox{0.9\columnwidth}{!}{%
%\vspace{0.2cm}
    \begin{tabular}{lc}
    \hline
    \multicolumn{1}{c}{\textbf{Parameter}} & \textbf{Value} \\ \hline
    Variant Type & \makecell{rand/1, best/1, rand/2, best/2, \\current-to-best/1, current-to-rand/1} ~\cite{georgioudakis2020comparative} \\ \hline
    Scaling Factor ($F$) & $0.50, 0.75, 1.00$ ~\cite{storn1997differential} \\ \hline
    \end{tabular}
}
\end{table}

Table~\ref{tab:param_de} reports the parameter settings we used for DE. We chose the six most widely used DE variants~\cite{georgioudakis2020comparative}. Regarding the scaling factor $F$, we chose values between 0.50 and 1.00 since 0.5 is a good initial choice and values below 0.4 or above 1.00 are usually not effective~\cite{storn1997differential}.
%We used the binary tournament for selecting individuals undergoing variation~\cite{goldberg1991comparative}, the blx-$\alpha$ for crossover~\cite{eshelman1993real}, random mutation~\cite{michalewicz1992genetic}, and both elitist and non-elitist methods for survival. Specifically, we tested three values for $\alpha$ in the crossover, which dictates how far the offspring are generated from the parents -- these values were selected based on a comparative study on self-adaptation among crossover methods~\cite{beyer2000desired} to determine the range of probability distribution between parents. Since the effects of mutation cannot be predicted on the specific problem (i.e., what happens with different degrees of exploration), we tested four different probabilities of mutation, including a strategy that dynamically and linearly decreases it from 1 to 0 with the generation counter (i.e., full exploration at early generations, full exploitation toward the end). The non-elitist survival method simply replaced the entire population with the offspring (i.e., stores the best solution without having it undergo the genetic operations), whereas the alternative elitist method employs the $\mu + \lambda$ schema~\cite{beyer2002evolution} (i.e., merges the old population and the offspring, sorts them by fitness with Rank Partitioning, and selects only the $N$ best individuals).

\subsubsection{Big Bang-Big Crunch Algorithm (BBBC)}
\label{sec:bbbc}
BBBC is designed for optimization in continuous domains by simulating the evolutionary process of the universe's expansion (Big Bang) and contraction (Big Crunch)~\cite{erol2006new}. In the Big Bang phase, candidate solutions, represented as points in a multi-dimensional space, are randomly scattered to explore the search space. During the Big Crunch phase, these points are drawn toward a central point (usually the best solution found so far) to exploit promising regions. This contraction of solutions guides the search directions, allowing the algorithm to refine and intensify the exploration around the most optimal areas. BBBC dynamically balances exploration and exploitation by iterating between these two phases, enhancing its ability to escape local optima and converge toward global optima, making it effective for solving complex, nonlinear optimization problems.

BBBC does not feature any parameter, so the only settings we used are the shared ones reported in Table~\ref{tab:param_shared}. 

\begin{figure*}[t!]
    \centering
    \subfigure[\protect\url{}\label{fig:best_task1}Task 1]
    {\includegraphics[height=6.6cm]{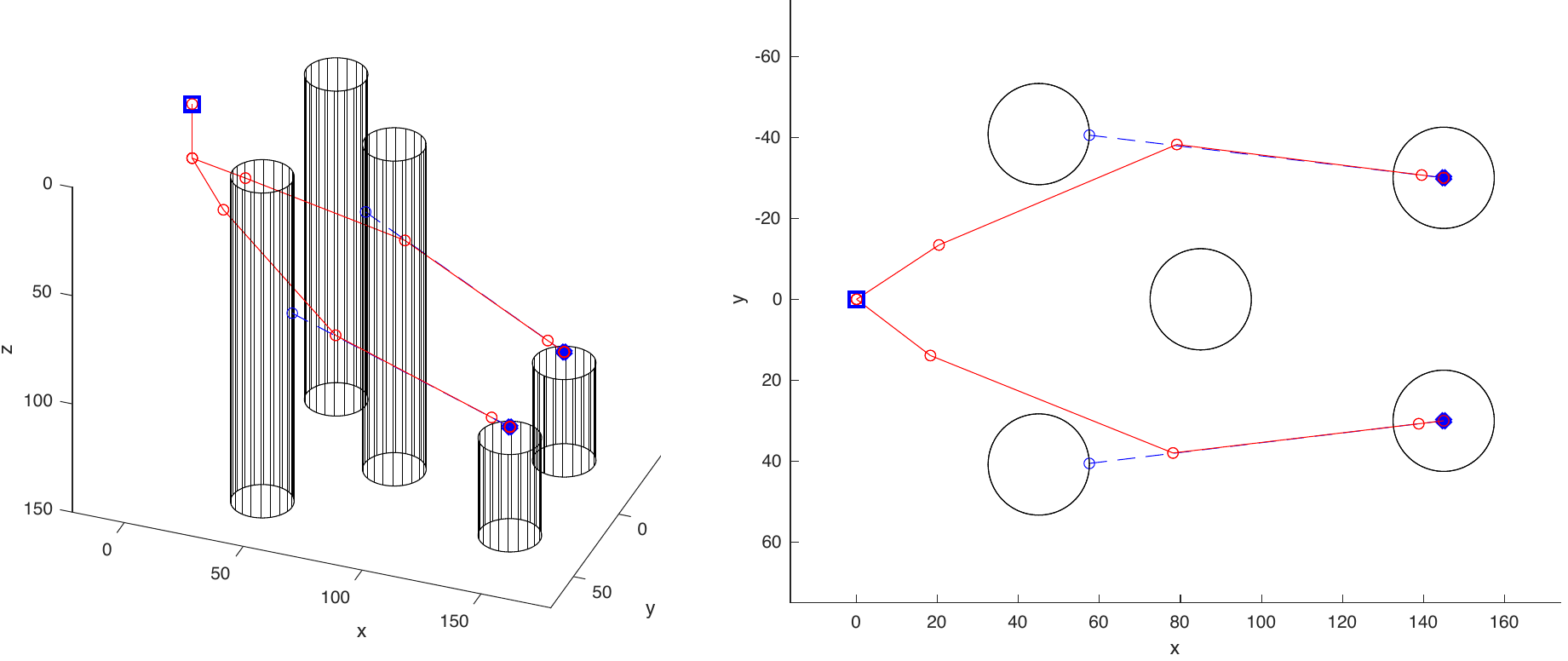}} 
    \subfigure[\protect\url{}\label{fig:best_task2}Task 2]
    {\includegraphics[height=6.6cm]{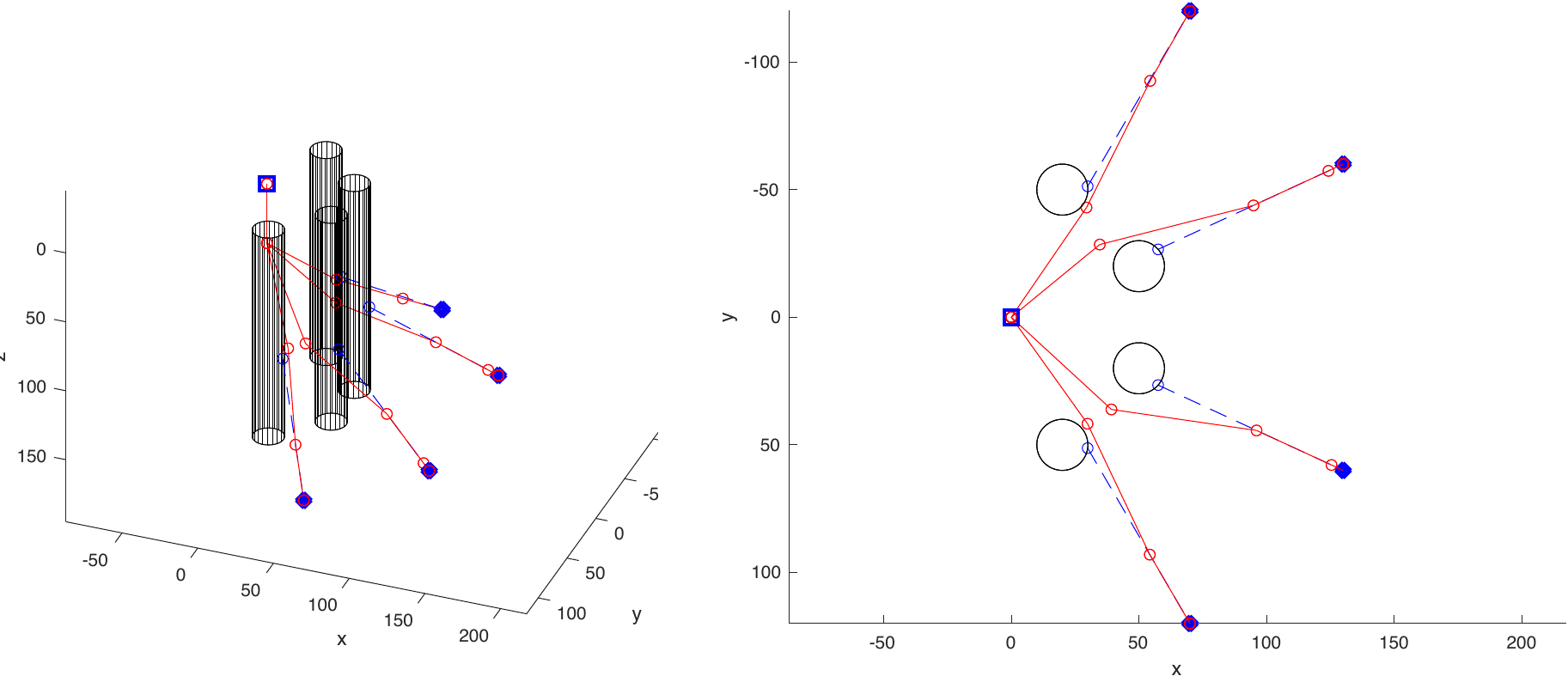}}
    \subfigure[\protect\url{}\label{fig:best_task3}Task 3]
    {\includegraphics[height=6.6cm]{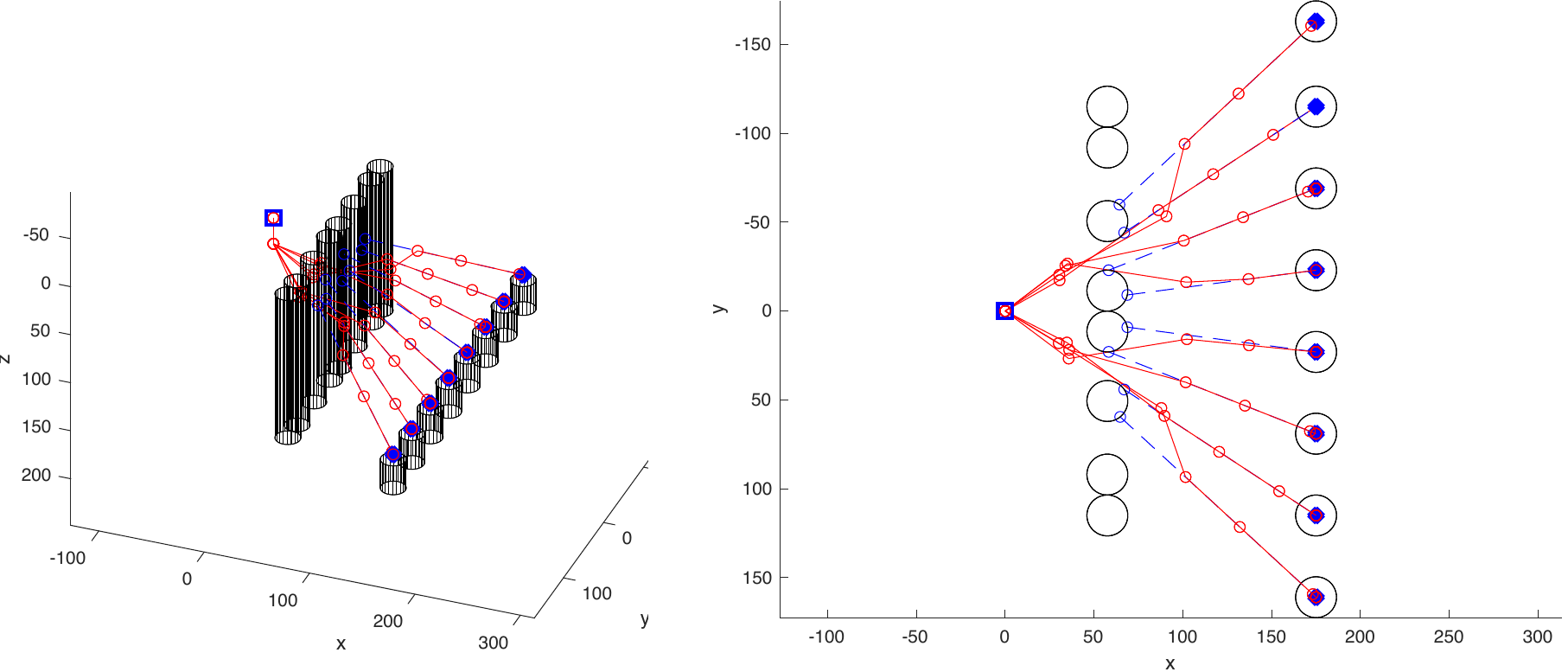}}
    \caption{Representation of each task in both isometric (left) and top view (right), along with their respective solutions.}
    \label{fig:optimal_solutions}
\end{figure*}

\subsection{Tasks}
\label{sec:tasks}

We selected three tasks with progressively increasing difficulty for which we could estimate that an optimal solution exists. In all tasks, the robot base is hung on the ceiling and reaches the targets from the top -- specifically, at 150 cm from the ground\footnote{The unit of measurement is irrelevant, as the problem can be scaled based on the size of the workspace; however, the dimensions we tried are based on realistic measurements in our environment, so we report them in cm.}. The tasks are:

\begin{itemize}
    \item \textbf{Task 1}: two targets placed on two pillars and three additional pillars as obstacles, a simple task in which the obstacles do not hinder the reachability of the targets (shown in Fig.~\ref{fig:best_task1});
    \item \textbf{Task 2}: four targets placed on the ground (i.e., a desk) and four pillars as obstacles, a medium task in which the obstacles are in the path of the targets (shown in Fig.~\ref{fig:best_task2}); and
    \item \textbf{Task 3}: eight targets placed on eight pillars and eight additional pillars as obstacles, a hard task in which the robot must find openings within this ``wall'' to reach many targets (shown in Fig.~\ref{fig:best_task3}).
\end{itemize}
We executed each task with a maximum link number $n=20$, joint rotation bounds set to $\Delta_{\theta}=\pi/4$ on each direction (a limited value emulating the physical robot's capabilities), and link length bounds set to $\Delta_{l} \in [25, 70]$ to fit the workspace. We repeated each task fifty times.

Concerning dimensionality, each rotational joint comprises two decision variables (i.e., rotation on $x$ and $y$) multiplied for each $|t|$ target (i.e., each configuration of the robot aims at one specific target in the environment). Configurations share link lengths, so they add a fixed constant value $n$. The overall dimensionality is $2 \cdot |t| \cdot n + n$, which is upper-bounded by $2 \cdot |t| \cdot n^{\Delta_\theta} + n^{\Delta_l}$. Specifically, the tasks we tested have the following dimensionalities:
\begin{itemize}
    \item Task 1: $2 \cdot 20 \cdot 2 + 20 = 100$ decision variables;
    \item Task 2: $2 \cdot 20 \cdot 4 + 20 = 180$ decision variables; and
    \item Task 3: $2 \cdot 20 \cdot 8 + 20 = 340$ decision variables.
\end{itemize}

\subsection{Parameter Combinations}
\label{sec:combinations}

The experiments present the following parameter combinations:

\begin{itemize}
    \item 3 tasks (see Section~\ref{sec:tasks});
    \item 4 optimization algorithms (see Section~\ref{sec:optimization_methods}), of which:
        \begin{itemize}
            %\item 24 combinations for GA: 3 values for blx-$\alpha$, 4 mutation probabilities, and 2 survival types (see Section~\ref{sec:ga} and Table~\ref{tab:param_ga});
            \item 12 combinations for GA: 3 values for blx-$\alpha$ and 4 mutation probabilities (see Section~\ref{sec:ga} and Table~\ref{tab:param_ga});
            \item 27 combinations for PSO: 3 values for inertia weight, 3 cognitive constants, and 3 social constants (see Section~\ref{sec:pso} and Table~\ref{tab:param_pso});
            \item 18 combinations for DE: 6 variants, and 3 scaling factors (see Section~\ref{sec:de} and Table~\ref{tab:param_de}); and
            \item 1 combination for BBBC (see Section~\ref{sec:bbbc});
        \end{itemize}
    %summing up to the 70 parameter combinations reported in Table S.1 of the Supplementary Material; and
    summing up to the 58 parameter combinations reported in Table S.1 of the Supplementary Material; and
    %\item 20 runs for each combination.
    \item 50 runs for each combination.
\end{itemize}
This sums up to $58 \cdot 50 = 2900$ runs for each task and $2900 \cdot 3 = 8700$ runs overall. 

\subsection{Evaluation Metric}
\label{sec:metrics}

Since we are dealing with a five-objective optimization problem (three objectives in which one of them is further divided into two, as described in Section~\ref{sec:fitness}), we used the rank from Rank Partitioning as the evaluation metric. The Rank Partitioning algorithm, described in Section 2 of the Supplementary Material, sorts the retrieved solutions based on their objective values according to their predefined priorities; then, it assigns a rank to each of them that linearly increments in ascending order from best to worst -- i.e., the best solution gets rank $=1$, the second best gets rank $=2$, and so on.

Therefore, we evaluated the results by collecting the $2900$ retrieved solutions for each task and assigned them a rank with Rank Partitioning -- i.e., from $1$ to $2900$. 

\subsection{Results}
\label{sec:results}

\begin{figure*}[t!]
    \centering
    \includegraphics[width=\linewidth]{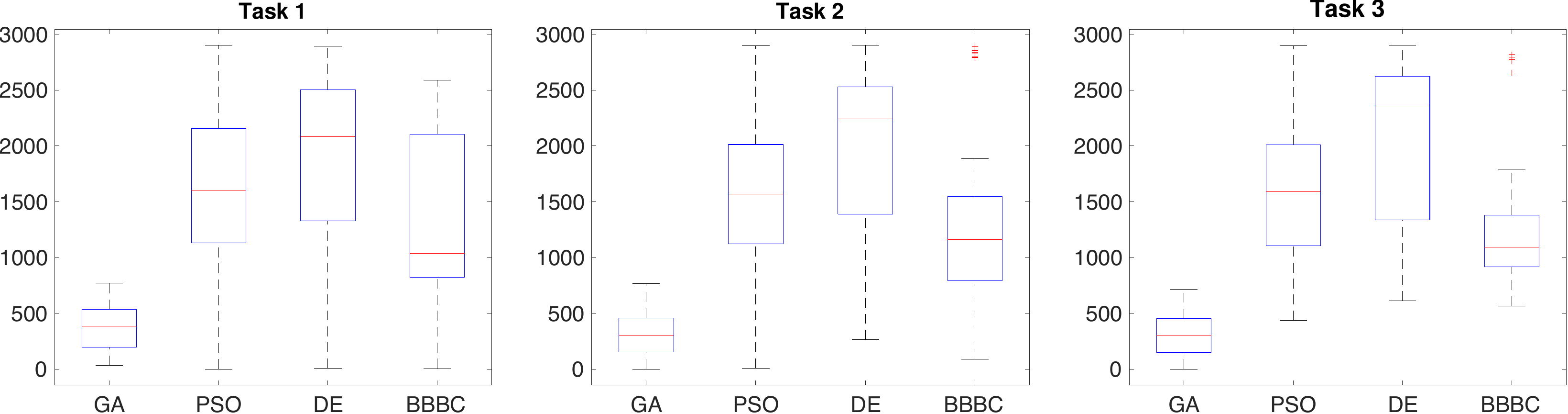}
    \caption{Ranking according to Rank Partitioning, over three tasks and each EC algorithm. Optimality is indicated by low-rank values (i.e., minimize rank). The same plot, divided among all parameter combinations among each EC algorithm, is shown in Fig. S.3 of the Supplementary Material.}
    \label{fig:boxplot_rankEverything}
\end{figure*}

Fig.~\ref{fig:boxplot_rankEverything} reports the rank results for each task and each EC algorithm, including median, interquartile range with outliers, and max/min across all runs. Separate results for each objective are reported in the Supplementary Material in Figure S.2, whereas a complete rank comparison among each of the 58 parameter combinations is depicted in Figure S.3. 
Table~\ref{tab:best_results} reports the numerical values of the best solution retrieved by each algorithm in each task, where the asterisk $^*$ indicates the solution at rank 1 depicted in Fig.~\ref{fig:optimal_solutions}.

% \usepackage{multirow}
% \usepackage{graphicx}
\begin{table}[]
\centering
\caption{Objective values of the best designs retrieved by each EA. The overall best (according to Rank Partitioning and depicted in Figure~\ref{fig:optimal_solutions}) is indicated with a $^*$.}
\label{tab:best_results}
\resizebox{\columnwidth}{!}{%
\begin{tabular}{cl|c|c|c|c|c|c|c|}
\cline{3-9}
\multicolumn{1}{l}{} &  & \textbf{IK ($f_{1-2}$)} & \textbf{LtS ($f_{3.1a}$)} & \textbf{UND ($f_{3.3}$)} & \textbf{LoS ($f_{3.1b}$)} & \textbf{LEN ($f_{3.2}$)} & \textbf{Rank} & \textbf{Constraints} \\ \hline
\multicolumn{1}{|c|}{\multirow{4}{*}{\rotatebox[origin=c]{90}{\textbf{Task 1}}}} & GA & $0.00$~cm & $6$ & $50\%$ & $2$  & $198.68$~cm & $32/2900$ & feasible \\
\multicolumn{1}{|c|}{} & PSO* & $0.25$~cm & $6$ & $25\%$  & $4$ & $197.57$~cm  & $1/2900$ & feasible \\
\multicolumn{1}{|c|}{} & DE & $0.43$~cm & $6$ & $25\%$  & $4$ & $196.97$~cm  & $8/2900$ & feasible \\
\multicolumn{1}{|c|}{} & BBBC & $0.37$~cm & $6$ & $25\%$  & $4$ & $197.90$~cm  & $5/2900$ & feasible \\ \hline
\multicolumn{1}{|c|}{\multirow{4}{*}{\rotatebox[origin=c]{90}{\textbf{Task 2}}}} & GA* & $0.26$~cm & $12$ & $31\%$ & $6$  & $224.87$~cm & $1/2900$ & feasible \\
\multicolumn{1}{|c|}{} & PSO & $0.50$~cm & $12$ & $38\%$ & $4$ & $226.49$~cm & $8/2900$ & feasible \\
\multicolumn{1}{|c|}{} & DE & $0.42$~cm & $12$ & $56\%$ & $5$ & $234.35$~cm & $264/2900$ & feasible \\
\multicolumn{1}{|c|}{} & BBBC & $0.42$~cm & $12$ & $44\%$ & $8$ & $221.81$~cm & $89/2900$ & feasible \\ \hline
\multicolumn{1}{|l|}{\multirow{4}{*}{\rotatebox[origin=c]{90}{\textbf{Task 3}}}} & GA* & $0.31$~cm & $27$ & $52\%$ & $21$  & $301.99$~cm & $1/2900$ & feasible \\
\multicolumn{1}{|l|}{} & PSO & $2.98$~cm & $28$ & $57\%$  & $22$ & $302.47$~cm & $436/2900$ & feasible \\
\multicolumn{1}{|l|}{} & DE & $12.96$~cm & $26$ & $54\%$  & $19$ & $295.20$~cm  & $612/2900$ & feasible \\
\multicolumn{1}{|l|}{} & BBBC & $7.99$~cm & $28$ & $45\%$  & $22$ & $291.71$~cm  & $565/2900$ & feasible \\ \hline
\end{tabular}%
}
\end{table}

We adopted Friedman's test with the Bonferroni-Dunn procedure~\cite{demvsar2006statistical} to test the robustness of each EC algorithm on the 58 parameter combinations for a total of 1653 pairwise comparisons (results are reported in the Supplementary Material under Tables S.2--S.5). 
The p-values show the result of the Bonferroni-Dunn procedure (ordered from Task 1 to 3 from top to bottom for each pairwise comparison); ``Y'' indicates that there is a significant difference (at the significance level $\alpha = 0.05$) between the corresponding pairwise settings, whereas ``N'' indicates no significant difference.

\subsection{Discussion}
\label{sec:discussion}

The optimal designs shown in Fig.~\ref{fig:optimal_solutions} (whose values are reported in Table~\ref{tab:best_results}) suggest that the optimizer is able to solve this complex engineering problem with no visible challenge. While Task 1 is easy and does not require specific tuning or a specific EC algorithm to be solved, the other tasks are more complex. Yet, all the designs can precisely reach the targets (precision is reported under $f_{1-2}$ in Table~\ref{tab:best_results}), show smooth and non-wavy configurations, with a small number of links (i.e., minimizing material waste). Furthermore, all the solutions are feasible, meaning that each algorithm was able to satisfy the constraints imposed in Section~\ref{sec:constraints}. The most surprising outcome, however, is that the configurations in a single design are almost mirrored. This was unexpected as EC algorithms are stochastic algorithms, and each configuration in the same design is not affected by the other in terms of joint actuation. We purposely chose mirrored tasks (i.e., all obstacles and targets are placed in the same fashion on the left-hand side as well as on the right-hand side) such that a mathematically optimal solution would retrieve a perfect mirrored solution. The top-view plots in Fig.~\ref{fig:optimal_solutions} show that our designs are very close to being mirrored, which is evidence of the optimality of our method: this was particularly unexpected in Task 3, and we could claim that no artificial fix-ups are required to smooth these configurations (i.e., manually remove the right-hand side configurations and replace them with a mirrored copy of the left-hand side ones).

In terms of EC algorithms, the numerical results show that GA is consistently more efficient in retrieving a better solution. This is visible from the plots in Fig.~\ref{fig:boxplot_rankEverything}: in all three tasks, the solutions from GA occupy the first ranks according to Rank Partitioning. This is even more visible on Task 3, which is the hardest of the three: while the other EC algorithms occasionally retrieve low-rank solutions in the first two tasks  (e.g., the best solution for Task 1 was retrieved with PSO as shown in Table~\ref{tab:best_results}), GA retrieved all the solutions from rank $1$ to $435$ in Task 3 -- meaning that the best $15\%$ solutions were all retrieved by GA -- whereas the worst algorithm for this type of problem is DE.

Specifically, according to the Friedman's test and Bonferroni-Dunn procedure (Tables S.2--S.5 of the Supplementary Material), there are no significant differences among all the GA settings, in all the tasks. This indicates that no specific parameter setting of Table~\ref{tab:param_ga} is more suitable than another, and designers can simply select a GA with any of these values. 
Furthermore, we found no statistical difference, in all tasks, between the following (with reference to the IDs in Table S.1 of the Supplementary Material): 
\begin{itemize}
    \item all GA settings and PSO 4 ($\omega = 0.6$, $c_1 = 2.5$, $c_2 = 0.5$);
    \item all GA settings and PSO 13 ($\omega = 0.8$, $c_1 = 2.5$, $c_2 = 0.5$);
    \item four out of twelve GA settings and DE 6 (best/1, $F = 1.0$); and
    \item three out of twelve GA settings and DE 15 (current-to-best/1, $F = 1.0$).
\end{itemize}
This indicates that GA, PSO 4, and PSO 13 are the best algorithms for our soft growing robot design.
Regarding BBBC, it was never found to be non-significantly different than any GA settings in all three tasks (for Task 2, GA was always significantly better). However, it was found to be non-significantly different than PSO 4, PSO 13, DE 6, and DE 15 (the settings that were non-significantly different than GA). By looking at the plots in Fig.~\ref{fig:boxplot_rankEverything}, we could claim that BBBC is, on average, more consistent than PSO and DE; although specific settings for the latter two often outperform it. This is interesting, considering that BBBC is a less-known metaheuristic, which yet could compete with big names in the state-of-the-art.

Lastly, we discuss the potential of the optimizer to solve problems with high dimensionality -- with reference to the values reported in Section~\ref{sec:tasks}. In the previous 2D work~\cite{stroppa2024design}, the algorithm was tested only up to 140 decision variables. The tasks analyzed in the current work are inherently more complex due to the third dimension of the environment; yet, as shown by the results reported in Section~\ref{sec:results}, the optimizer can tackle high dimensionality problems while still achieving (sub)optimal designs -- especially when the core EC algorithm is GA.

In conclusion, we found that our optimizer can retrieve high-performance 3D designs in terms of precision in reaching targets, resource consumption, and actuation, especially when the core EC algorithm is GA.

\section{Conclusion}
\label{sec:conclusion}

This work presents an extension of our tool for soft-growing design optimization~\cite{stroppa2024design}, offered to robotic enthusiasts when facing a specific task. While the previous work was limited only to planar manipulators (i.e., in 2D), the current work completes the tool addressing more realistic 3D scenarios. We used the Rank Partitioning algorithm, introduced in our previous work, to solve a multi-objective optimization problem without compromising among trade-offs. Since this algorithm can theoretically work on every population-based optimizer, we tested it on four different Evolutionary Computation techniques: Genetic Algorithm, Particle Swarm Optimization, Differential Evolution, and Big Bang-Big Crunch algorithm. 
Results show that the proposed optimizer retains the outstanding performances of its previous 2D version, especially when based on a Genetic Algorithm.

In future works, we will use the optimizer on a real specific task, manufacture the soft robot according to the retrieved design, and use our motion planner~\cite{altagiuri2024motion} to demonstrate the robot's capabilities as a manipulator. However, this poses two new challenges that could not be addressed in this work: (i) a proper control module is required to address not only kinematic actuation but also dynamic components, and (ii) the amount of material loss in the kinematic chain when links bend, which is a real-case problem that was not modeled in this work. Additionally, we plan to include a further objective in the optimization problem: minimize the path between configurations such that the robot visits all targets with the optimized order -- also known as Traveling Salesman Problem~\cite{hoffman2013traveling}. This objective cannot be handled directly in the optimization due to its extreme computational complexity; however, it is useless to calculate the optimal order for solutions far away from the optimum (i.e., high-rank solutions according to Rank Partitioning). Therefore, this objective could be evaluated only for a subset of the individuals in the population having low rank, reducing the amount of computation required~\cite{kulz2024optimizing}.

\section*{Acknowledgments}

This work is funded by TUBİTAK within the scope of the 2232-B International Fellowship for Early Stage Researchers Program number 121C145.

The authors would like to thank: 
Prof. Allison M. Okamura for inspiring the project,
Kadir Kaan Atalay and Kemal Erdem Yenin for writing the code of BBBC, 
Kerem Bicen for writing the code of DE, 
%\todo{Ayodele Oyejide for preparing the CAD of the soft growing robot},
Ayodele Oyejide for providing pictures of a real soft growing robot,
and Emre Ozel for allowing us to use 121 computers in the Kadir Has University's Computer Labs for parallel computation.

\section*{Supplementary Information}

A document containing the Supplementary Material can be downloaded at \url{https://www.mediafire.com/file/s4r3yqm489cccg6/3D_SGR_supplementary_material.pdf/file}.

%The MATLAB script is fully available at \url{https://www.mediafire.com/file/pwvzb1u6nw4k9e5/soft-robots-studio-3d.zip/file}.

A graphical presentation on the Rank Partitioning algorithm, including theory and hands calculation, is available at \url{https://www.mediafire.com/file/6jlp1yr4p8yjjul/22._Rank_Partitioning_for_MOEAs.pdf/file}.

Readers interested in manufacturing these soft growing robots can find more information at \url{https://www.vinerobots.org/}.

\bibliographystyle{elsarticle-num} 
\bibliography{references}

\end{document}